\documentclass{article}

 \usepackage[main, final]{neurips_2026}

\usepackage[utf8]{inputenc} 
\usepackage[T1]{fontenc}    
\usepackage{hyperref}       
\usepackage{url}            
\usepackage{booktabs}       
\usepackage{amsfonts}       
\usepackage{nicefrac}       
\usepackage{microtype}      
\usepackage{xcolor}         
\usepackage{enumitem}
\usepackage{amsmath}
\usepackage{makecell}
\usepackage{soul}
\usepackage[normalem]{ulem}
\usepackage{multirow}
\usepackage{multicol}
\usepackage{subcaption}
\usepackage{bm}
\usepackage{graphicx}
\usepackage[most]{tcolorbox} 
\title{Beyond Words: Multimodal LLM Knows\\When to Speak}

\author{%
  Zikai Liao$^{1}$ \hspace{1mm} Yi Ouyang$^{2}$ \hspace{1mm} Yi-Lun Lee$^{2}$ \hspace{1mm} Chen-Ping Yu$^{2}$ \hspace{1mm} Yi-Hsuan Tsai$^{2}$ \hspace{1mm} Zhaozheng Yin$^{1}$ \\
  Department of Computer Science, Stony Brook University$^{1}$\\
  Atmee AI$^{2}$\\
}

\begin{document}

\maketitle

\vspace{-1em}
\begin{abstract}
    Chatbots via large language models (LLMs) generate fluent responses but often struggle with when to speak, especially for brief, timely listener reactions during ongoing dialogue. We present a multimodal strategy for LLMs, which leverages synchronized video, audio, and text cues to improve conversational timing awareness. The strategy reformulates response timing as a dense response-type prediction task, enabling an agent to decide whether to remain silent, produce a short reaction, or start a full response under streaming constraints. Therefore, we introduce a curated multimodal dataset from real-world dyadic conversational videos with temporally aligned modalities and fine-grained reaction type annotations. Moreover, we design a multimodal strategy, MM-When2Speak, with a multimodal integration module on top of an LLM backbone. Experiments across various modality settings and strong LLM baselines show that MM-When2Speak achieves up to a 3$\times$ improvement in response type prediction performance, highlighting the importance of multimodal perception for natural and engaging conversational interaction.
\end{abstract}
\vspace{-1em}
\section{Introduction} \label{sec:1}

Recent large language model (LLM)-based assistants have demonstrated strong capabilities in generating coherent and contextually appropriate responses. 
However, natural interaction depends not only on \textit{what to say}, but also on \textit{when to speak}, particularly for brief listener-like reactions (e.g., acknowledgments, agreement, short questions) that occur during an ongoing partner utterance. 
In human conversations, speakers rely on contextual cues, social norms, intentions, and subtle verbal or bodily signals to decide whether to remain silent, offer a short backchannel, or initiate a response \cite{wang2024nips,ekstedt2020turngpt,stivers2009universals,umair2024llmknows,wittenburg2006elan,yngve1970getting}. 
Accurately predicting these moments is crucial for enabling fluid and engaging human–machine interaction.

Despite their success, many LLMs \cite{achiam2023gpt,liu2024deepseekv3,ekstedt2020turngpt,ekstedt2023rcturngpt} operate in structured, turn-based settings and lack explicit mechanisms for timing prediction, limiting their ability to capture the dynamism of natural dialogue \citep{wang2024nips}. 
Prior work on backchannel detection \cite{amer2023backchannel,wang2023unveiling,park2024backchannel,inoue2025yeah,fukunaga2025backchannel,cieri2004fisher,bilakhia2015mahnob,wu2020harpervalleybank,lee2025behavior} partially addresses this issue, yet often overlooks the diversity and contextual variability of reaction types, constraining nuanced conversational behavior.

Enabling timely and socially appropriate responsiveness remains challenging due to limited multimodal integration. 
Two key issues arise: 
\textbf{(1) Dataset limitations.} Most existing resources are text-based \cite{umair2024llmknows,pilan2024conversational,mahowald2024dissociating,cieri2004fisher} or audio-centric \cite{bilakhia2015mahnob,lee2025behavior}, while visual signals such as head movements, gaze, and facial affect are underrepresented despite their importance in face-to-face and video-mediated dialogue. 
\textbf{(2) Modeling limitations.} Many LLM systems follow a text-centric pipeline, disregarding prosodic and visual cues that humans naturally exploit to anticipate turn transitions and produce timely reactions. 
Consequently, current systems often lack the contextual grounding necessary for real-time, socially aligned interaction.

In this work, we argue that \textbf{equipping LLMs with multimodal perception provides a practical and transferable strategy} for response timing in conversational agents.
We adopt a lightweight yet general recipe: (i) represent conversational context with synchronized video, speech, and text streams; (ii) reformulate ``when to speak'' as a dense temporal decision problem; and (iii) adaptively fuse multimodal cues to predict response or reaction types at fine temporal granularity. 
This strategy can function as a plug-in controller for various LLMs, deciding whether to remain silent, produce a short reaction, or initiate a full response.

To support this study, we curate a multimodal dyadic conversational dataset with temporally aligned visual, auditory, and textual streams. 
Beyond coarse turn-switching labels (\texttt{full\_response}, \texttt{silence}), we annotate nuanced reaction categories including \texttt{affirmation}, \texttt{gratitude}, \texttt{farewell}, \texttt{greeting}, \texttt{question}, \texttt{surprise}, and \texttt{pondering}, extending ``when to speak'' with amore fine-grained modeling. 
This design enables systematic evaluation of response timing and reaction type selection under realistic conversational dynamics.

\begin{figure}[t]
    \centering
    \includegraphics[width=\linewidth,keepaspectratio]{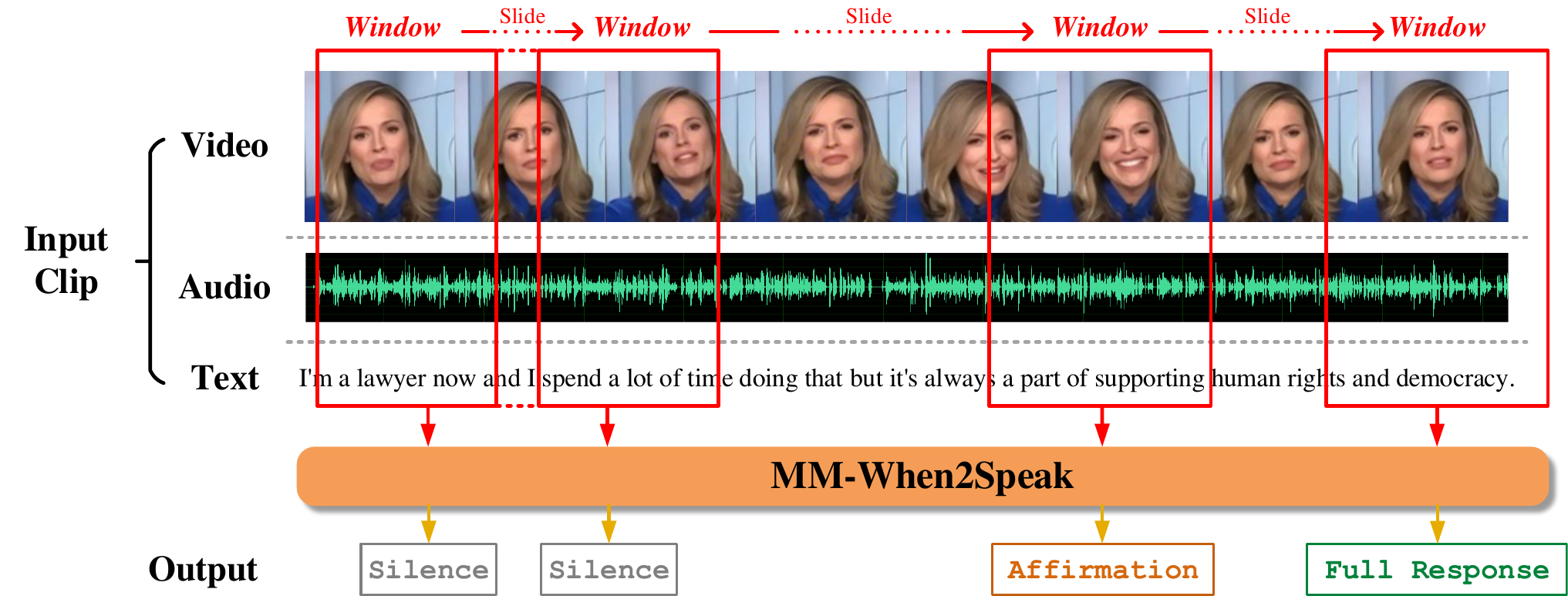}
    \vspace{-1em}
    \caption{Overview of MM-When2Speak. It uses a sliding window to densely sample short clips for response type prediction, transforming ``when to speak'' problem to a classification task. At each sampled timestamp, it outputs a specific label, indicating whether to keep silent, give a short reaction (e.g., \texttt{Affirmation}), or start responding.}
    \vspace{-1.5em}
    \label{fig:overview}
\end{figure}

We instantiate this strategy in MM-When2Speak (Fig.~\ref{fig:overview}), which predicts when to respond, react, or remain silent. 
The system integrates visual, auditory, and textual signals through a lightweight fusion module and formulates the task as multi-class classification. 
A sliding-window mechanism enables online inference for real-time deployment. 
Extensive comparisons against strong LLM-based baselines under varying modality settings show up to a 3$\times$ improvement in response type prediction, demonstrating the critical role of multimodal input in achieving accurate and natural conversational behavior. Our main contributions are summarized as follows:

\vspace{-0.5em}
\begin{itemize}[leftmargin=1.em, itemsep=0em, topsep=0em]
    \item 
    We curate a dyadic video conversation dataset with synchronized visual, speech, and transcript streams, annotated with both turn transitions and fine-grained reaction categories, enabling systematic study of multimodally grounded response timing in realistic conversational scenarios.

    \item 
    We propose a practical and transferable recipe that equips LLM-based conversational agents with aligned visual, auditory, and textual perception, reformulating ``when to speak'' as a prediction problem with multimodal cues.

    \item 
    We instantiate the proposed strategy in a multimodal classification framework with online inference capability, and demonstrate up to a 3$\times$ improvement over strong LLM-based baselines under varying modality settings, highlighting the critical role of visual and prosodic signals in natural conversational dynamics.
\end{itemize}
\vspace{-1em}
\section{Related Work}

\vspace{-0.5em}
\subsection{Response Timing and Backchannel Detection}



Response timing and backchannel detection are central to human-like conversational modeling \cite{umair2024llmknows}. Early dialogue systems relied on fixed silence thresholds to detect turn shifts, often leading to delayed or overlapping responses \cite{umair2024llmknows,wang2024nips}. Recent transformer-based approaches model transition relevance places (TRPs) and backchannel opportunities more explicitly. For example, \cite{umair2024llmknows} shows that modern LLMs “know what to say but not when to speak,” highlighting their difficulty in detecting within-utterance openings and proposing new TRP benchmarks. \cite{wang2024nips} introduces a full-duplex LLM-based dialogue framework that emits control tokens to decide whether to listen, interrupt, or respond. However, these approaches primarily operate on textual signals and do not explicitly model listener backchannel behaviors.

Beyond textual modeling, multimodal approaches have been explored for backchannel prediction, including joint listener–speaker architectures \cite{fukunaga2025backchannel} and multi-party conversational datasets such as MPIIGroupInteraction \cite{muller2021multimediate,muller2022multimediate,muller2023multimediate,muller2024multimediate}. These works focus on predicting engagement or listener feedback but typically do not jointly address response timing and backchannel generation in dyadic conversations. 

In contrast, our MM-When2Speak framework jointly models response timing and listener backchannel behaviors, enabling systems to determine both \textit{when} to respond and \textit{how} to react in conversational interactions.

\vspace{-0.5em}
\subsection{Multimodal Large Language Models}



Multimodal large language models (MLLMs) extend LLMs beyond text-only reasoning by integrating visual, video, and speech modalities. Early works such as Flamingo \cite{alayrac2022flamingo} and BLIP-2 \cite{li2023blip2} align frozen LLMs with visual encoders through lightweight adapters, enabling few-shot image understanding and multimodal dialogue. Subsequent efforts extend these models to video–language tasks \cite{liu2023visual,maaz2024videochatgpt,wang2024videollm}, supporting temporal grounding and real-time video–text interaction. In the speech domain, AudioPaLM \cite{rubenstein2023audiopalm} integrates PaLM-2 \cite{anil2023palm} with AudioLM \cite{borsos2023audiolm} for unified speech understanding and generation.

While these models demonstrate strong multimodal perception and reasoning capabilities, they mainly focus on understanding or generating multimodal content rather than modeling conversational dynamics. In contrast, we leverage multimodal conversational cues to model response timing and equip LLMs with the ability to predict \textit{when to speak} during human interactions.
\vspace{-0.8em}
\section{Methodology}

\vspace{-0.5em}
\subsection{Problem Formulation} \label{sec:3.1}

We consider a dyadic interaction between a \textit{user} and a \textit{machine}, where the machine must decide at each moment whether to (i) remain silent, (ii) produce a brief reaction, or (iii) take the conversational turn and deliver a full response. 
We formulate this as a response-type classification task over short conversational clips. 
Each clip is assigned a response type label:

\vspace{-1em}
{\scriptsize
\[
\hat{y} \in \{\texttt{affirmation}, \texttt{gratitude}, \texttt{farewell}, \texttt{greeting}, \texttt{question}, \texttt{surprise}, \texttt{pondering}, \texttt{full\_response}, \texttt{silence}\},
\]
}

\noindent where the first seven categories correspond to brief listener reactions that do not shift the conversational floor, \texttt{full\_response} indicates a turn-taking action, and \texttt{silence} denotes no response.

Given a full video $\mathcal{V} = (V_{\mathcal{V}}, S_{\mathcal{V}}, T_{\mathcal{V}})$ containing synchronized visual, audio, and transcript streams, we define that a response type prediction at time $t_i$ is based on a previous multimodal context window of duration $\Delta t$, which is a short clip $C_i$ sampled from the video defined as:
\[
C_i = \text{Crop}(\mathcal{V}, t_i - \Delta t, t_i) = (V_{C_i}, S_{C_i}, T_{C_i}),
\]
where $V_{C_i}$, $S_{C_i}$, and $T_{C_i}$ denote the video frames, audio features, and transcript tokens within the context window.
Then the model processes the short clip and outputs a response type:
\[
\hat{y}_i = \text{Cls}(C_i),
\]
which produces a sequence $\{\hat{y}_i\}_{i=1}^{N}$ over $N$ sampled clips. 
For each prediction at time $t_i$: if $\hat{y}_i = \texttt{silence}$, the machine should remain quiet; if $\hat{y}_i = \texttt{full\_response}$, it shall initiate a turn; otherwise, it produces a brief reaction of the specified type without taking the floor.

\vspace{-0.5em}
\subsection{Collection and Curation of Our Dataset} \label{sec:3.2}

We collect over 2,000 dyadic conversation videos from public online platforms (e.g., YouTube), each featuring two individuals engaged in frontal-face, split-screen dialogues in various settings such as online meetings and broadcast news.
To ensure data quality, we apply a series of filtering steps to remove low-quality samples, excluding videos with poor or inconsistent face visibility, excessive background noise unrelated to the conversation, or substantial overlapping speech that hinders reliable transcription. After this pre-processing pipeline, we retain 377 high-quality, full-length videos for further annotation.

We then obtain valid short clips by using audio diarization \cite{pyannote2024hub} to separate speakers and break down the composition of the conversations in each video, which enables to locate context clip based on diarized timestamps. Specifically, a short clip is marked as a \texttt{full\_response} context if the opposite participant starts to speak after the given context; it is labeled as a \texttt{reaction} if a brief utterance (not \texttt{full\_response}) is produced right afterwards; otherwise, the clip will be assigned as \texttt{silence}. To further classify the type of each \texttt{reaction} clip, we provide its text transcript to ChatGPT and categorize the responses into seven fine-grained reaction types as described in Sec.~\ref{sec:3.1}. 

We first collect short clips from 357 videos among the entire 377 high-quality videos. We construct the \textbf{Short-Clips} dataset by stratifiedly sampling 4,393 \texttt{reaction}, 2,000 \texttt{full\_response}, and 2,000 \texttt{silence} segments. This dataset is divided into training and testing subsets by a ratio of 7:3, denoted as \textbf{Short-Clips-Train} for model training and \textbf{Short-Clips-Test} for evaluation, respectively. 
The dataset is specifically designed to assess a model’s ability to predict appropriate response types from isolated short clips.

Second, we use the remaining 20 videos to construct the \textbf{Full-Videos} dataset for evaluating the model’s ability to determine ``when to speak'' in a continuous dialogue setting. For each video, we obtain its annotations by labeling overlapping short clips with a duration $\Delta t$ and a stride $\delta$ from the very beginning, resulting in a dense sequence of time-aligned response-type annotations that reflect the ground truth temporal decisions. This setup enables fine-grained temporal evaluations in real-world scenarios.

To verify the annotation quality, we conduct a human evaluation study on the Short-Clips dataset, as described in the following Sec. \ref{sec:humanstudy}. More details on the data collection pipeline, segment pre-processing, response type annotation, dataset construction, and uncertainty handling for full-video setup are provided in Appendix \ref{sec:x-1}.

\vspace{-0.5em}
\subsection{MM-When2Speak} \label{sec:3.3}

\begin{figure}[t]
    \centering
    \includegraphics[width=\linewidth,keepaspectratio]{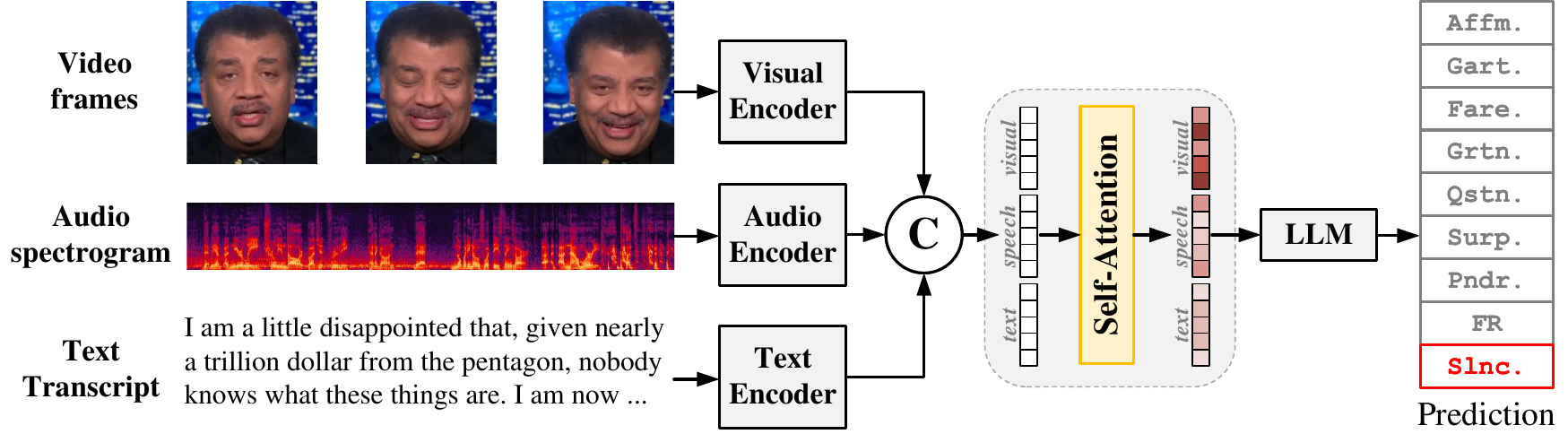}
    \vspace{-1.5em}
    \caption{Architecture of MM-When2Speak. It encodes visual frames ($V_{C_i}$), audio spectrogram features ($S_{C_i}$), and tokenized texts ($T_{C_i}$) into modality-specific representations, then adaptively fuses them for the LLM backbone to predict response types\protect\footnotemark.}
    \vspace{-1.5em}
    \label{fig:architecture}
\end{figure}

\footnotetext{
    \fontsize{7.5pt}{6pt}\selectfont
    {\texttt{Affm.}=\texttt{affirmation},\quad\texttt{Grat.}=\texttt{gratitude},\quad\texttt{Fare.}=\texttt{farewell},\quad\texttt{Grtn.}=\texttt{greeting},\quad\texttt{Qstn.}=\texttt{question},\quad\texttt{Surp.}=\texttt{surprise},\quad\texttt{Pndr.}=\texttt{pondering},\quad\texttt{FR.}=\texttt{full\_response},\quad\texttt{Slnc.}=\texttt{silence}.
    }
}

\textbf{Multimodal fusion strategy.} Our core strategy is to equip an LLM with multimodal perception by (i) extracting modality-specific representations, (ii) projecting them into a shared embedding space, and (iii) performing adaptive fusion so the model can emphasize the most informative cues for response-type prediction.
Concretely, given a clip $C_i=(V_{C_i}, S_{C_i}, T_{C_i})$, we obtain visual, audio, and textual embeddings and concatenate them into a single token sequence. We then apply self-attention over the concatenated tokens to perform modality-adaptive aggregation, which allows the model to selectively attend to salient signals (e.g., facial dynamics or prosody) and remain robust under modality imbalance. A high-level overview is shown in Fig.~\ref{fig:architecture}, with details in Appendix \ref{sec:x-2}.

We instantiate the above strategy with an ``Encoder--Adaptor--LLM'' pipeline following \cite{fu2025vita}. For vision, we adopt InternViT \cite{chen2024internvl} as the visual encoder. For audio, we convert raw waveforms into Mel spectrogram features and encode them using convolutional downsampling followed by Transformer blocks, following \cite{wang2024freeze}. We then choose Qwen2-7B-base \citep{yang2024qwen2} as the LLM backbone, with a hidden size  of 4,096. Modality-specific features are mapped into the LLM embedding space via lightweight MLP adaptors before fusion and classification.

\textbf{Training procedure.} We adopt a two-stage training scheme inspired by \cite{fu2025vita}. We first perform multimodal pretraining to encourage cross-modal consistency among visual, audio, and textual representations, which helps stabilize fusion and improve robustness. We then fine-tune the full model on Short-Clips-Train with supervised response type classification. To mitigate class imbalance across reaction categories, we use Focal Loss with $\gamma=2$ and $\alpha=0.5$. More implementation details are provided in Appendix \ref{sec:x-2}.
\vspace{-1em}
\section{Experiments}

\begin{table}[t]
    \fontsize{8pt}{6.5pt}\selectfont
    \setcellgapes{1.5pt}
    \makegapedcells
    \setlength{\tabcolsep}{2pt} 
    \centering
    \caption{Performance evaluations of response type prediction on Short-Clips-Test dataset for methods of different modalities.}
    \begin{tabular}{c|c|cccccccccccccccccc}
        \Xhline{1.5pt}
        Method & Metric & \texttt{\texttt{Affm.}} && \texttt{\texttt{Grat.}} && \texttt{\texttt{Fare.}} && \texttt{\texttt{Grtn.}} && \texttt{\texttt{Qstn.}} && \texttt{\texttt{Surp.}} && \texttt{\texttt{Pndr.}} && \texttt{FR} && \texttt{\texttt{Slnc.}} \\
        \hline\hline
        \multicolumn{19}{c}{\textit{Text}} \\
        \hline
        \multirow{2}{*}{ChatGPT-5.2} & P & 22.51 && 19.63 && 21.54 && 20.50 && 18.04 && 17.29 && 17.12 && 22.98 && 28.55 \\
        & R & 18.47 && 15.65 && 17.53 && 16.51 && 14.10 && 13.37 && 13.20 && 18.94 && 24.44 \\ \hline
        \multirow{2}{*}{DeepSeek-V3} & P & 13.74 && 19.90 && 20.63 && 22.39 && 12.85 && 12.26 && 12.32 && 20.25 && 21.91 \\
        & R & 10.56 && 16.60 && 17.32 && 19.06 && 9.69 && 9.12 && 9.18 && 16.94 && 18.58 \\ \hline
        \multirow{2}{*}{Gemini-3.1}  & P & 21.91 && 21.63 && 25.59 && 17.49 && 13.30 && 14.08 && 16.99 && 22.21 && 26.15 \\
        & R & 18.16 && 17.88 && 21.80 && 13.82 && 9.75 && 10.50 && 13.33 && 18.45 && 22.34 \\ \hline
        \multirow{2}{*}{Qwen-2.5}   & P & 19.74 && 16.76 && 14.37 && 13.05 && 13.03 && 10.76 && 14.39 && 20.64 && 21.88 \\
        & R & 16.44 && 13.51 && 11.17 && 9.89 && 9.87 && 7.68 && 11.19 && 17.33 && 18.55 \\ \hline
        \multirow{2}{*}{VITA-1.5}  & P & 15.32 && 17.08 && 15.19 && 14.04 && 14.78 && 12.68 && 13.34 && 20.46 && 21.49 \\
        & R & 12.10 && 13.82 && 11.97 && 10.85 && 11.57 && 9.53 && 10.17 && 17.15 && 18.17\\
        \hline\hline
        \multicolumn{19}{c}{\textit{Audio + Text}} \\
        \hline
        \multirow{2}{*}{ChatGPT-5.2} & P & 26.85 && 20.76 && 27.76 && 23.68 && 21.07 && 21.57 && 20.70 && 27.05 && 31.43 \\
        & R & 24.19 && 18.14 && 25.09 && 21.04 && 18.44 && 18.94 && 18.08 && 24.39 && 28.75\\ \hline
        \multirow{2}{*}{Gemini-3.1}  & P & 30.49 && 30.38 && 26.24 && 24.25 && 18.06 && 19.21 && 19.30 && 28.08 && 28.84 \\
        & R & 27.81 && 27.70 && 23.58 && 21.60 && 15.46 && 16.60 && 16.69 && 25.41 && 26.17\\ \hline
        \multirow{2}{*}{Qwen2-Audio} & P & 29.49 && 24.83 && 21.09 && 19.02 && 18.76 && 19.00 && 15.43 && 25.13 && 30.22 \\
        & R & 26.82 && 22.18 && 18.46 && 16.41 && 16.15 && 16.39 && 12.86 && 22.48 && 27.54\\ \hline        
        \multirow{2}{*}{VITA-1.5}  & P & 24.64 && 20.34 && 26.03 && 25.59 && 16.86 && 24.24 && 16.97 && 30.70 && 32.69 \\
        & R & 21.99 && 17.72 && 23.37 && 22.94 && 14.27 && 21.59 && 14.38 && 28.02 && 30.00\\
        \hline\hline
        \multicolumn{19}{c}{\textit{Video + Text}} \\
        \hline
        \multirow{2}{*}{ChatGPT-5.2} & P & 33.16 && 29.38 && 33.59 && 35.61 && 28.66 && 29.31 && 26.88 && 33.81 && 39.42 \\
        & R & 30.79 && 27.02 && 31.22 && 33.23 && 26.30 && 26.95 && 24.53 && 31.44 && 37.04 \\ \hline
        \multirow{2}{*}{Gemini-3.1} & P & 33.77 && 35.40 && 31.66 && 28.47 && 23.52 && 25.37 && 27.37 && 29.50 && 34.27 \\
        & R & 31.57 && 33.19 && 29.46 && 26.28 && 21.34 && 23.19 && 25.18 && 27.30 && 32.96 \\ \hline
        \multirow{2}{*}{Qwen2.5-VL} & P & 21.89 && 23.87 && 20.32 && 19.69 && 19.30 && 20.88 && 21.19 && 23.78 && 24.52 \\
        & R & 17.65 && 19.83 && 17.25 && 17.14 && 16.01 && 15.42 && 16.98 && 20.09 && 19.49 \\ \hline
        \multirow{2}{*}{VITA-1.5}  & P & 25.04 && 24.23 && 20.88 && 26.81 && 20.69 && 25.09 && 18.56 && 33.17 && 34.96 \\
        & R & 23.12 && 22.31 && 18.97 && 24.88 && 18.78 && 23.17 && 16.66 && 31.23 && 33.02\\
        \hline\hline
        \multicolumn{19}{c}{\textit{Video + Audio + Text}} \\
        \hline
        \multirow{2}{*}{VITA-1.5} & P & 38.03 && 39.68 && 39.16 && 38.81 && 28.46 && 31.23 && 23.32 && 31.36 && 35.32 \\
        & R & 35.14 && 36.79 && 36.27 && 35.92 && 25.61 && 28.37 && 20.50 && 28.50 && 32.44\\ \hline
        \multirow{2}{*}{MM-When2Speak} & P & \textbf{62.21} && \textbf{64.35} && \textbf{63.15} && \textbf{63.29} && \textbf{46.26} && \textbf{50.52} && \textbf{37.78} && \textbf{68.15} && \textbf{68.78} \\
        & R & \textbf{59.86} && \textbf{61.99} && \textbf{60.79} && \textbf{60.44} && \textbf{43.91} && \textbf{46.25} && \textbf{35.45} && \textbf{65.79} && \textbf{66.42}\\
        \Xhline{1.5pt}
    \end{tabular}
    \vspace{-2em}
    \label{tab:short-clips}
\end{table}

\vspace{-0.5em}
\subsection{Experimental Setup}  \label{sec:4.1}

We evaluate the response-type prediction performance of MM-When2Speak against a range of state-of-the-art models under various modality configurations, in which we use task-specific prompts to constrain model outputs for predefined response labels. The full prompt template we use is provided in Appendix \ref{sec:x-3}.

First, we compare our method with ChatGPT-5.2-Thinking \cite{openaiGPT52}, DeepSeek-V3-7B \cite{liu2024deepseekv3}, Qwen-2.5-7B \cite{yang2024qwen25}, Gemini-3.1-Pro \cite{deepmindGemini31} and VITA-1.5 \cite{fu2025vita} for the \textit{Text} setting. For multimodal settings, we include ChatGPT-5.2-Thinking, Qwen2-Audio \cite{chu2024qwen2audio}, Gemini-3.1-Pro and VITA-1.5; for \textit{Audio+Text}, while for \textit{Video+Text}, we include ChatGPT-5.2-Thinking, Gemini-3.1-Pro, Qwen2.5-VL \cite{bai2025qwen2.5-vl}, and VITA-1.5. Finally, in the full \textit{Video+Audio+Text} setting, we compare against VITA-1.5, one of the few public models supporting all three modalities.

For proprietary models (ChatGPT-5.2-Thinking, Gemini-3.1-Pro), inference is conducted via official APIs, while other models are deployed locally on a single NVIDIA L40s.Comparison methods are evaluated in a zero-shot manner.
We use precision (P), recall (R), and F1 score metrics to evaluate the response-type prediction performance of each method.

\subsection{Comparisons with State-of-the-art LLMs}
\vspace{-0.5em}

We evaluate on both datasets: Short-Clips-Test measures response-type classification from isolated multimodal snippets, while Full-Videos evaluates dense, time-aligned predictions in continuous real-world dialogues.

\textbf{Short-Clips.} As shown in Table~\ref{tab:short-clips}, incorporating additional modalities consistently improves response-type prediction for all LLM baselines. Our MM-When2Speak achieves the best overall performance under the full \textit{Video+Audio+\\Text} setting, outperforming text-only LLMs by a large margin and remaining noticeably stronger than the strongest multimodal baseline (up to 3$\times$ over ChatGPT-5.2-Thinking on \textit{Text}, and 1.5$\times$ over VITA-1.5 on \textit{Video+Audio+Text}).

\begin{table}[t]
    \fontsize{8pt}{6.5pt}\selectfont
    \setcellgapes{1.5pt}
    \makegapedcells
    \setlength{\tabcolsep}{2pt} 
    \centering
    \caption{Performance evaluations of response type prediction on Full-Videos dataset for methods of different modalities.}
    \begin{tabular}{c|c|cccccccccccccccccc}
        \Xhline{1.5pt}
        Method & Metric & \texttt{\texttt{Affm.}} && \texttt{\texttt{Grat.}} && \texttt{\texttt{Fare.}} && \texttt{\texttt{Grtn.}} && \texttt{\texttt{Qstn.}} && \texttt{\texttt{Surp.}} && \texttt{\texttt{Pndr.}} && \texttt{FR} && \texttt{\texttt{Slnc.}} \\
        \hline\hline
        \multicolumn{19}{c}{\textit{Text}} \\
        \hline
        \multirow{2}{*}{ChatGPT-5.2} & P & 11.50 && 9.36 && 8.58 && 8.66 && 8.53 && 9.32 && 8.89 && 11.64 && 12.22 \\
        & R & 8.71 && 6.65 && 5.89 && 5.98 && 5.85 && 6.60 && 6.20 && 8.84 && 9.41 \\ \hline
        \multirow{2}{*}{DeepSeek-V3} & P & 9.46 && 7.69 && 6.82 && 8.34 && 7.69 && 7.97 && 8.72 && 10.39 && 10.95 \\
        & R & 6.87 && 5.18 && 4.36 && 5.80 && 5.18 && 5.45 && 6.16 && 7.77 && 8.31 \\ \hline
        \multirow{2}{*}{Gemini-3.1}   & P & 10.19 && 8.53 && 8.74 && 8.23 && 8.70 && 7.84 && 8.68 && 9.56 && 11.37 \\
        & R & 7.53 && 5.94 && 6.14 && 5.65 && 6.10 && 5.28 && 6.08 && 6.92 && 8.68 \\ \hline
        \multirow{2}{*}{Qwen-2.5}   & P & 11.29 && 7.55 && 9.78 && 8.92 && 9.17 && 9.59 && 7.95 && 10.47 && 10.69 \\
        & R & 8.64 && 5.05 && 7.18 && 6.35 && 6.59 && 7.00 && 5.43 && 7.85 && 8.06 \\ \hline
        \multirow{2}{*}{VITA-1.5}  & P & 10.38 && 9.64 && 9.47 && 7.95 && 8.52 && 8.39 && 9.25 && 10.53 && 11.18 \\
        & R & 7.76 && 7.04 && 6.88 && 5.43 && 5.97 && 5.85 && 6.67 && 7.90 && 8.53 \\ 
        \hline\hline
        \multicolumn{19}{c}{\textit{Audio + Text}} \\
        \hline
        \multirow{2}{*}{ChatGPT-5.2} & P & 14.15 && 12.50 && 15.54 && 13.86 && 16.31 && 14.42 && 16.19 && 15.38 && 13.22 \\
        & R & 11.64 && 10.02 && 12.99 && 11.37 && 13.68 && 11.90 && 13.65 && 12.85 && 10.73 \\ \hline
        \multirow{2}{*}{Gemini-3.1} & P & 13.90 && 13.38 && 14.25 && 13.62 && 14.43 && 14.15 && 15.18 && 13.22 && 13.27 \\
        & R & 11.49 && 10.97 && 11.78 && 11.20 && 12.01 && 11.72 && 12.74 && 10.81 && 10.86 \\ \hline
        \multirow{2}{*}{Qwen2-Audio}   & P & 11.99 && 12.77 && 13.29 && 13.86 && 15.19 && 14.57 && 13.21 && 14.26 && 11.80 \\
        & R & 9.64 && 10.41 && 10.92 && 11.48 && 12.79 && 12.18 && 10.84 && 11.88 && 9.46 \\ \hline
        \multirow{2}{*}{VITA-1.5}  & P & 12.97 && 12.62 && 12.75 && 11.53 && 15.09 && 15.33 && 14.07 && 15.33 && 13.60 \\
        & R & 10.61 && 10.26 && 10.39 && 9.19 && 12.70 && 12.93 && 11.69 && 12.93 && 11.23 \\ 
        \hline\hline
        \multicolumn{19}{c}{\textit{Video + Text}} \\
        \hline
        \multirow{2}{*}{ChatGPT-5.2} & P & 19.95 && 18.01 && 20.90 && 20.44 && 19.58 && 21.39 && 19.83 && 22.46 && 21.45 \\
        & R & 17.00 && 15.08 && 17.94 && 17.49 && 16.64 && 18.43 && 16.88 && 19.49 && 18.50 \\ \hline
        \multirow{2}{*}{Gemini-3.1} & P & 19.77 && 18.01 && 18.41 && 18.25 && 20.83 && 17.86 && 19.25 && 19.94 && 19.56 \\
        & R & 16.93 && 15.19 && 15.38 && 15.27 && 17.98 && 15.05 && 16.41 && 17.09 && 16.73 \\ \hline
        \multirow{2}{*}{Qwen2.5-VL}   & P & 15.44 && 17.67 && 15.54 && 18.03 && 13.70 && 15.25 && 13.64 && 18.44 && 18.38 \\
        & R & 18.58 && 16.03 && 17.12 && 14.52 && 13.95 && 17.25 && 17.51 && 15.12 && 16.88 \\ \hline
        \multirow{2}{*}{VITA-1.5}  & P & 20.06 && 17.44 && 18.76 && 17.48 && 18.51 && 18.27 && 18.94 && 21.24 && 21.15 \\
        & R & 17.27 && 14.68 && 15.98 && 14.72 && 15.73 && 15.50 && 16.16 && 18.44 && 18.35 \\ 
        \hline\hline
        \multicolumn{19}{c}{\textit{Video + Audio + Text}} \\
        \hline
        \multirow{2}{*}{VITA-1.5}  & P & 21.40 && 18.94 && 22.18 && 17.73 && 21.97 && 22.15 && 20.72 && 26.21 && 27.05 \\
        & R & 18.95 && 16.51 && 19.72 && 15.31 && 19.52 && 19.67 && 18.27 && 23.73 && 24.57 \\ \hline
        \multirow{2}{*}{MM-When2Speak}  & P & \textbf{31.55} && \textbf{32.26} && \textbf{31.25} && \textbf{29.53} && \textbf{28.21} && \textbf{28.82} && \textbf{27.47} && \textbf{35.17} && \textbf{33.27} \\
        & R & \textbf{29.24} && \textbf{29.95} && \textbf{28.94} && \textbf{27.22} && 25.91 && \textbf{26.52} && \textbf{25.17} && \textbf{32.85} && \textbf{30.95} \\         
        \Xhline{1.5pt}
    \end{tabular}
    \vspace{-2em}
    \label{tab:full-videos}
\end{table}


\textbf{Full-Videos.} Table~\ref{tab:full-videos} shows a similar trend: multimodal inputs yield higher accuracy, and MM-When2Speak performs the best overall under dense sliding-window inference. Absolute performance is lower than Short-Clips, as Full-Videos reflects more realistic conversational dynamics and introduces more challenging class imbalance (especially \texttt{silence}) and ambiguous boundary cases under our formulation.


\textbf{Latency.} We report inference latency in Table~\ref{tab:latency}. While latency naturally increases with more modalities, MM-When2Speak remains comparable to representative baselines under each modality configuration, indicating practical feasibility for real-time deployment. 

We also conduct case studies for representative samples in our dataset compared with other methods. Due to space limits, we include the results in Appendix \ref{sec:x-4}.

\begin{table}[t]
    \fontsize{8pt}{6.5pt}\selectfont
    \setcellgapes{1.5pt}
    \makegapedcells
    \setlength{\tabcolsep}{6pt} 
    \centering
    \caption{Latency comparison of different models across modalities in seconds.}
    \begin{tabular}{l|c|c|c|c|c|c}
        \Xhline{1.2pt}
        Modality & \multicolumn{2}{c|}{\textit{Text}} & \multicolumn{2}{c|}{\textit{Video+Text}} & \multicolumn{2}{c}{\textit{Video+Audio+Text}}  \\
        \hline
        Models & Qwen-2.5 & Ours & Qwen2.5-VL & Ours & VITA-1.5 & Ours \\
        \hline
        \hline
        Average & 0.103 & 0.092 & 0.561 & 0.606 & 1.132 & 1.145 \\
        Min. & 0.068 & 0.073 & 0.217 & 0.235 & 0.779 & 0.698 \\
        Max. & 0.388 & 0.369 & 0.835 & 0.825 & 1.889 & 1.732 \\
        \Xhline{1.2pt}
    \end{tabular}
    \label{tab:latency}
    \vspace{-1em}
\end{table}

\subsection{Ablation Studies and More Analyses}

We report two representative ablation studies and analysis in this section. Additional results, such as finetuned models comparison, evaluation on other datasets, robustness to noisy inputs, comparisons with backchannel detection method, transferability for multimodal strategy, and evaluations of different prompt templates, are included in Appendix \ref{sec:x-3} and \ref{sec:appendix-additional-ablations}.

\begin{table}[t]
    \fontsize{8pt}{6.5pt}\selectfont
    \setcellgapes{1.5pt}
    \makegapedcells
    \setlength{\tabcolsep}{3pt} 
    \centering
    \caption{Mulltimodal effectiveness of our MM-When2Speak on Short-Clips-Test.}
    \begin{tabular}{c|cccccccccccccccccc}
        \Xhline{1.5pt}
        Metric & \texttt{\texttt{Affm.}} && \texttt{\texttt{Grat.}} && \texttt{\texttt{Fare.}} && \texttt{\texttt{Grtn.}} && \texttt{\texttt{Qstn.}} && \texttt{\texttt{Surp.}} && \texttt{\texttt{Pndr.}} && \texttt{FR} && \texttt{\texttt{Slnc.}} \\
        \hline
        \multicolumn{18}{c}{(a) Ours (\textit{Video + Audio + Text})} \\
        \hline
        P   & 62.21 && 64.35 && 63.15 && 63.29 && 46.26 && 50.52 && 37.78 && 68.15 && 68.78 \\
        R   & 59.86 && 61.99 && 60.79 && 60.44 && 43.91 && 46.25 && 35.45 && 65.79 && 66.42 \\
        F1  & 61.01 && 63.15 && 61.95 && 62.09 && 45.05 && 49.62 && 36.58 && 66.95 && 67.58 \\
        \hline
        \multicolumn{18}{c}{(b) Ours w/o Self-attention (\textit{Video + Audio + Text})} \\
        \hline
        P   & 60.51 && 63.16 && 62.03 && 61.76 && 45.16 && 46.91 && 36.91 && 65.44 && 68.72 \\
        R   & 56.83 && 59.47 && 58.64 && 58.07 && 41.51 && 45.56 && 35.11 && 61.75 && 65.01 \\
        F1  & 58.61 && 61.26 && 60.43 && 59.86 && 43.26 && 47.71 && 35.01 && 63.54 && 66.82 \\
        \hline
        \hline
        \multicolumn{18}{c}{(c) Ours (\textit{Video + Text})} \\
        \hline
        P   & 36.89 && 38.65 && 41.09 && 38.83 && 34.22 && 32.71 && 31.75 && 43.49 && 42.32 \\
        R   & 44.22 && 44.41 && 40.45 && 42.93 && 36.07 && 33.72 && 24.12 && 46.05 && 41.49 \\
        F1  & 40.23 && 41.64 && 40.85 && 40.94 && 33.66 && 33.22 && 27.72 && 44.71 && 41.89 \\
        \hline
        \multicolumn{18}{c}{(d) Ours w/o Self-attention (\textit{Video + Text})} \\
        \hline
        P   & 35.98 && 37.69 && 40.06 && 37.86 && 33.37 && 31.91 && 30.96 && 42.41 && 41.27 \\
        R   & 43.12 && 43.30 && 39.50 && 41.87 && 35.17 && 31.91 && 23.52 && 45.91 && 40.40 \\
        F1  & 39.23 && 40.60 && 39.83 && 39.92 && 32.82 && 31.91 && 26.41 && 44.01 && 40.85 \\
        \hline
        \hline
        \multicolumn{18}{c}{(e) Ours (\textit{Audio + Text})} \\
        \hline
        P   & 29.96 && 31.39 && 33.36 && 31.53 && 27.79 && 26.56 && 25.78 && 36.54 && 34.27 \\
        R   & 34.59 && 34.73 && 31.68 && 33.58 && 25.91 && 25.95 && 28.85 && 32.74 && 32.40 \\
        F1  & 32.36 && 32.55 && 31.54 && 31.61 && 25.98 && 26.24 && 27.16 && 34.52 && 33.31 \\
        \hline
        \multicolumn{18}{c}{(f) Ours w/o Self-attention (\textit{Audio + Text})} \\
        \hline
        P   & 29.59 && 31.00 && 32.96 && 31.10 && 26.45 && 26.24 && 25.47 && 34.69 && 33.37 \\
        R   & 34.16 && 32.75 && 31.29 && 31.37 && 25.27 && 25.47 && 28.39 && 32.56 && 32.40 \\
        F1  & 32.28 && 31.47 && 32.05 && 31.23 && 25.87 && 25.84 && 26.82 && 33.55 && 32.87 \\
        \Xhline{1.5pt}
    \end{tabular}
    \vspace{-1em}
    \label{tab:ablation1}
\end{table}

\textbf{Multimodality and self-attention fusion}. Table~\ref{tab:ablation1} shows that removing any modality leads to clear performance drops, with the largest degradation occurring when visual cues are absent, indicating that video, speech, and text provide complementary signals for robust response-type prediction. Moreover, replacing self-attention fusion with a simpler alternative consistently hurts performance, validating the importance of attention-based multimodal aggregation.

\begin{figure}[t]
    \centering
    \includegraphics[width=\linewidth]{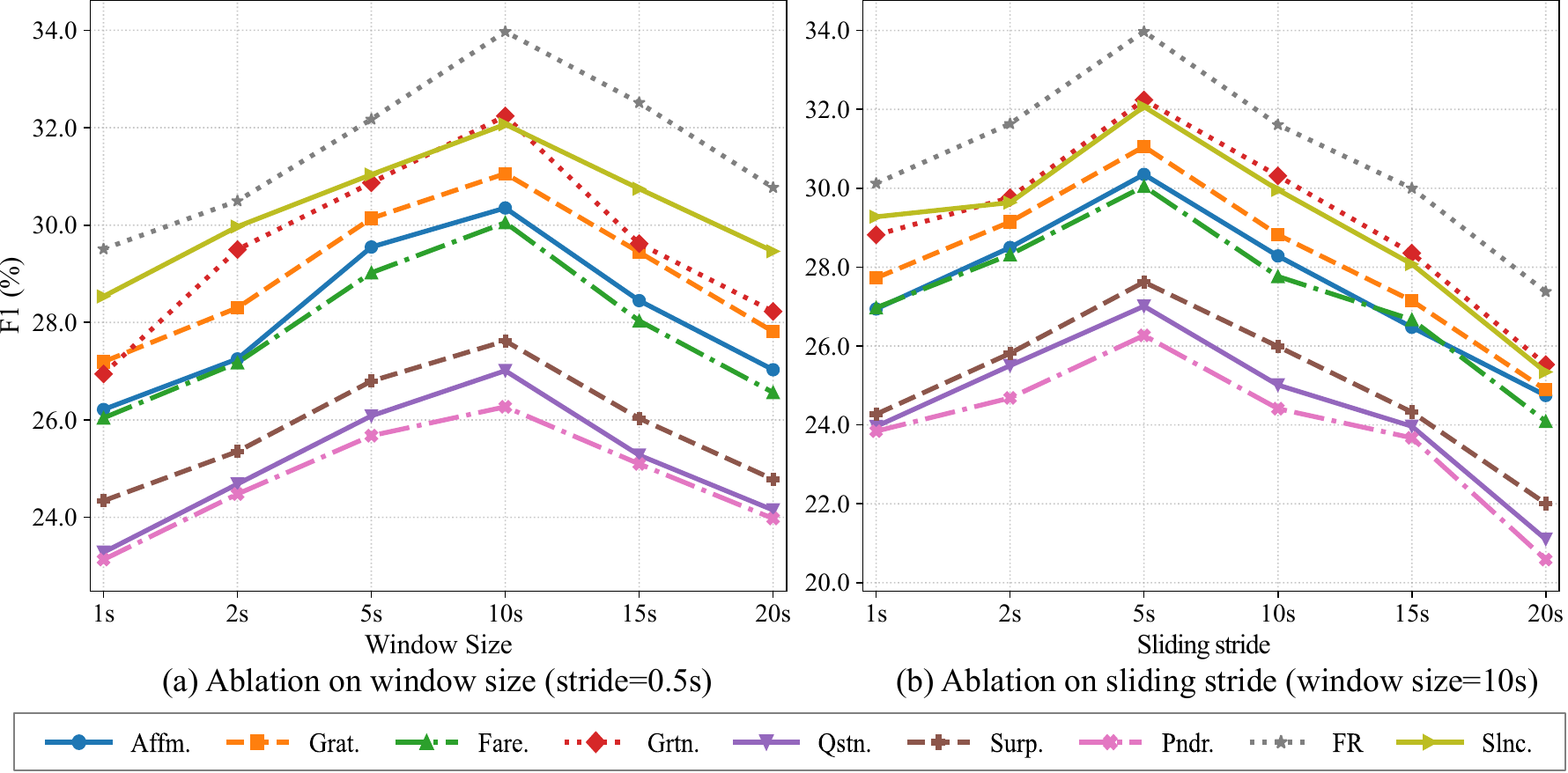}
    \vspace{-1.5em}
    \caption{Analysis on different window sizes and strides for the sliding window mechanism for Full-Videos evaluation.}
    \vspace{-1.5em}
    \label{fig:ablation-window-size-stride}
\end{figure}

\textbf{Size and stride of sliding window}.
We perform full-video inference using a sliding window mechanism with a window size of 10 seconds and a stride of 0.5 seconds. We ablate these two parameters to measure their impact on performance using the Full-Videos dataset, where we plot the results to display the trend in Fig. \ref{fig:ablation-window-size-stride}.
It can be observed that performance peaks when opting window size=10s and stride=0.5s, which verifies our design. This suggests that using sliding window with appropriate size and stride may include sufficient contexts for accurate response predictions.

\subsection{Human Study on Dataset Annotation Quality}  \label{sec:humanstudy}

As described in Sec.~\ref{sec:3.1}, we use ChatGPT to assign fine-grained reaction types for \texttt{reaction} segments. To validate the reliability of this automatic labeling, we conduct a human study on the Short-Clips dataset.

\textbf{Human Study Workflow}. We sample 105 clips (10s each, with video, audio, and transcript) using stratified sampling over classes and domains, consisting of 27 Affirmations, 2 Farewells, 6 Gratitudes, 3 Greetings, 6 Ponderings, 4 Questions, 6 Surprises, 21 Full Responses, and 30 Silences. Each of the nine volunteers is independently assigned with one of the nine response-type labels, using the same written class definitions as the pipeline; annotators are blinded to pipeline outputs and to each other. We do not enforce consensus or adjudication, as the goal is to measure alignment between the pipeline and a diverse set of human judgments. The information for the nine voluntary participants are reported in Table \ref{tab:participant-info}.

\vspace{-1em}
\begin{table}[t]
    \centering
    \fontsize{8pt}{6.5pt}\selectfont
    \setcellgapes{1.5pt}
    \makegapedcells
    \setlength{\tabcolsep}{2pt} 
    \caption{Basic information for participants in the human study.}
    \begin{tabular}{l|l}
        \Xhline{1.2pt}
        Item & Value \\
        \hline
        Number of participants & 9 \\
        Experience & 7/9 with prior multimedia labeling; all instructed with the procedure. \\
        Language & All fluent in English \\
        Training & 10--15 min guideline session + 10 practice items \\
        Compensation & Voluntary \\
        Ethics & No data distribution or re-identification; local offline use only \\
        Tooling & Google Forms /\; on-site labeling \\
        \Xhline{1.2pt}
    \end{tabular}
    \vspace{-1.5em}
    \label{tab:participant-info}
\end{table}

\vspace{-0.5em}
\begin{table}[!htbp]
    \fontsize{8pt}{6.5pt}\selectfont
    \setcellgapes{1.5pt}
    \makegapedcells
    \setlength{\tabcolsep}{2pt} 
    \centering
    \caption{Auto vs.\ Majority-Human Agreement.}
    \begin{tabular}{c|c|c|c|c|c}
        \Xhline{1.2pt}
         & Accuracy & Macro-Precision & Macro-Recall & Macro-F1 & CI (95\%) \\
        \hline
        Value & 0.876 & 0.881 & 0.851 & 0.865 & [0.834, 0.882] \\
        \Xhline{1.2pt}
    \end{tabular}
    \vspace{-1em}
    \label{tab:agreement}
\end{table}

\vspace{-1em}
\begin{table}[!htbp]
    \fontsize{7.5pt}{6pt}\selectfont
    \setcellgapes{1.5pt}
    \makegapedcells
    \setlength{\tabcolsep}{4pt} 
    \centering
    \caption{Per-class Metrics (Auto vs.\ Majority-Human).}
    \begin{tabular}{c|c|c|c|c|c|c|c|c|c|c}
        \Xhline{1.2pt}
        Metric & \texttt{Affm.} & \texttt{Grat.} & \texttt{Fare.} & \texttt{Grtn.} & \texttt{Qstn.} & \texttt{Surp.} & \texttt{Pndr.} & \texttt{FR} & \texttt{Slnc.} & Macro-Avg \\
        \hline
        P & 0.91 & 0.80 & 0.78 & 0.78 & 0.71 & 0.80 & 0.69 & 0.92 & 0.94 & --- \\
        R & 0.87 & 0.50 & 0.75 & 0.67 & 0.67 & 0.75 & 0.63 & 0.89 & 0.90 & --- \\
        F1 & 0.89 & 0.62 & 0.76 & 0.72 & 0.69 & 0.77 & 0.66 & 0.90 & 0.92 & 0.86 \\
        \Xhline{1.2pt}
    \end{tabular}
    \vspace{-0.5em}
    \label{tab:perclass-agreement}
\end{table}

\textbf{Auto vs. Majority-Human Agreement}. Table~\ref{tab:agreement} reports accuracy, macro precision/recall/F1, and a 95\% bootstrap CI for our automatic labels against Majority-Human labels, where higher values indicate stronger alignment with human judgments. All metrics exceed 0.85, supporting the reliability of our annotation pipeline. Using the same Majority-Human reference, Table~\ref{tab:perclass-agreement} further breaks down per-class precision/recall/F1. Agreement is highest on frequent categories (e.g., \texttt{Affm.}, \texttt{FR}, \texttt{Slnc.}) and shows expected variability on lower-support reaction types, consistent with the human study in Sec.~\ref{sec:humanstudy}.

\begin{figure}[t]
    \centering
    \includegraphics[width=\linewidth,keepaspectratio]{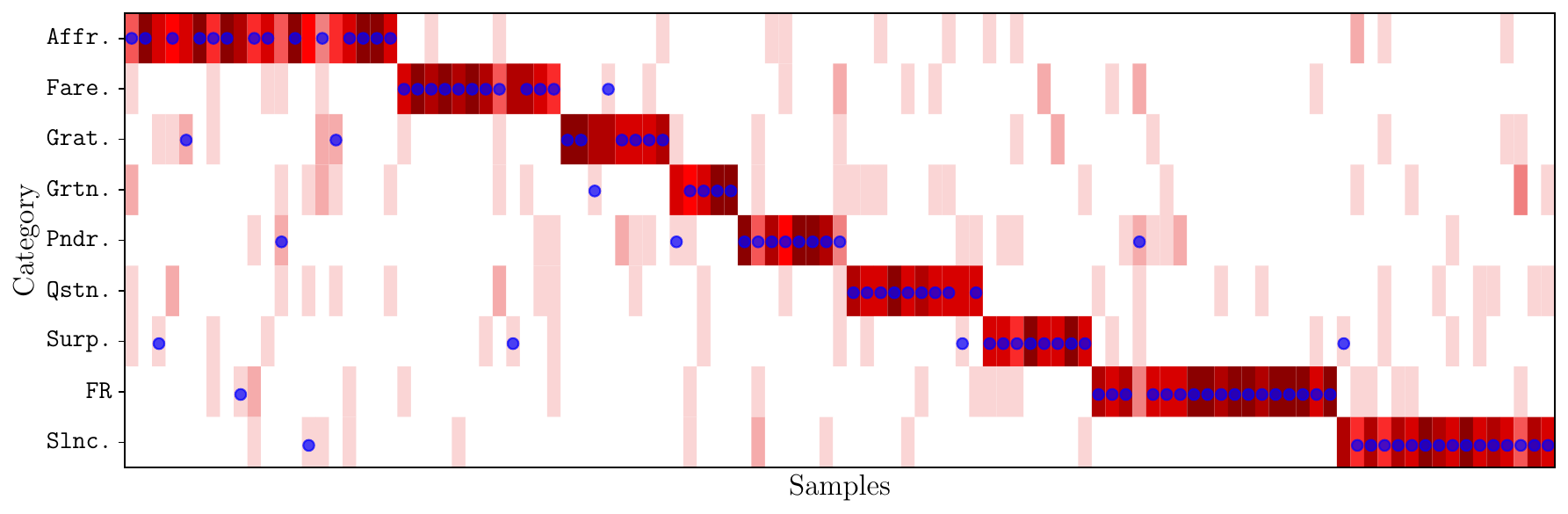}
    \vspace{-2em}
    \caption{Comparison between pipeline-generated and human labels. Red heatmaps show the distribution of human votes, while blue circles denote the pipeline labels. We observe 87.62\% agreement (92/105), supporting annotation reliability.}
    \vspace{-1em}
    \label{fig:humanstudy}
\end{figure}

We compare pipeline labels against the majority human vote. As shown in Fig.~\ref{fig:humanstudy}, the pipeline label matches the top-voted human category for 92 out of 105 samples (87.62\%), indicating strong agreement with human judgments.

\begin{table}[t]
    \fontsize{8pt}{6.5pt}\selectfont
    \setcellgapes{1.5pt}
    \makegapedcells
    \setlength{\tabcolsep}{4pt} 
    \centering
    \caption{Inter-annotator agreement by class (Fleiss' $\kappa$).}
    \begin{tabular}{c|c|c|c|c|c|c|c|c|c|c}
        \Xhline{1.2pt}
         & \texttt{Affm.} & \texttt{Grat.} & \texttt{Fare.} & \texttt{Grtn.} & \texttt{Qstn.} & \texttt{Surp.} & \texttt{Pndr.} & \texttt{FR} & \texttt{Slnc.} & Overall \\
        \hline
        $\kappa$ & 0.78 & 0.49 & 0.66 & 0.52 & 0.53 & 0.61 & 0.55 & 0.80 & 0.85 & 0.71 \\
        \Xhline{1.2pt}
    \end{tabular}
    \vspace{-1em}
    \label{tab:kappa}
\end{table}

\textbf{Inter-Annotator Agreement by Class}. Table~\ref{tab:kappa} reports inter-annotator agreement for each class. For each reaction type we convert the labels into a \textit{binary one-vs-rest scheme} (1 = this class, 0 = other), and compute Fleiss' $\kappa$ on the resulting matrix. The overall Fleiss' $\kappa\approx 0.71$ indicates substantial agreement, confirming that the reaction labels are sufficiently reliable to validate our automatic pipeline.

\vspace{-1em}
\begin{table}[!thbp]
    \fontsize{8pt}{6.5pt}\selectfont
    \setcellgapes{1.5pt}
    \makegapedcells
    \setlength{\tabcolsep}{2pt} 
    \centering
    \caption{Top-5 confusion pairs (percentage of 105 samples).}
    \begin{tabular}{c|c}
        \Xhline{1.2pt}
        True $\rightarrow$ Pred & Percentage (\%) \\
        \hline
        Question $\rightarrow$ Surprise & 5.8 \\
        Pondering $\rightarrow$ Silence & 3.9 \\
        Affirmation $\rightarrow$ Question & 2.4 \\
        Pondering $\rightarrow$ Affirmation & 2.9 \\
        Gratitude $\rightarrow$ Greeting & 1.9 \\
        \Xhline{1.2pt}
    \end{tabular}
    \vspace{-0.5em}
    \label{tab:confusions}
\end{table}

\textbf{Top-5 Confusions and Failure Cases}. Using Majority-Human labels as reference, we compute a confusion-frequency matrix between human annotations and automatic outputs. The five most frequent (out of 105 samples) are given in Table~\ref{tab:confusions}. These confusions mainly occur between semantically adjacent or subtle categories, consistent with human disagreement patterns. Most remaining errors arise from weak prosody, micro-expressions, or window-boundary effects:

\vspace{-0.5em}
\begin{itemize}[leftmargin=1.em, itemsep=0em, topsep=0em]
    \item \textbf{Pondering vs.\ Surprise.} 
    Both reactions have weak prosodic and lexical cues and rely on subtle facial signals; short $\tau=250$ ms windows may capture only partial reaction onset, increasing confusion.

    \item \textbf{Affirmation vs.\ Silence.} 
    Some very short backchannels (e.g., ``Right'', ``Yeah'') often have low energy or ambiguous features similar to silence or background noises, leading to silence predictions.

    \item \textbf{Greeting / Farewell / Question ambiguity.} 
    These short, context-dependent utterances already show lower human agreement ($\kappa\approx0.49$ -- $0.61$), suggesting part of the confusion reflects intrinsic annotation ambiguity.

    \item \textbf{Semantic proximity of discourse acts.} 
    Several confusion pairs correspond to conversational functions with overlapping cues (e.g., rising intonation in \textit{Question} and \textit{Surprise}), making class boundaries naturally soft.
\end{itemize}
\vspace{-1em}
\section{Conclusion}


In this work, we present MM-When2Speak with our proposed multimodal strategy for enabling LLM-based conversational agents to predict ``when to speak'' in human conversations. By reformulating response timing as a dense, window-based response-type classification problem, MM-When2Speak leverages synchronized multimodal signals to capture fine-grained conversational cues that are crucial for natural interaction, especially for subtle ``when to react'' behaviors. To support this study, we curate two complementary datasets and evaluation settings that cover both short-clip prediction and continuous full-video inference. Extensive experiments show that multimodal inputs consistently improve timing prediction for strong LLM baselines, and MM-When2Speak achieves the best overall performance across modalities and datasets.
In future work, we will scale to larger and more diverse conversational data, extend the framework to richer multi-party settings, and explore broader real-world scenarios to enable more complex conversational dynamics.

{
\small
\bibliographystyle{plain}
\bibliography{main}
}


\clearpage
\newpage
\appendix
\section*{\textbf{Appendix}}
The following is the contentof the supplementary material for our main paper. 

\begin{itemize}[leftmargin=1.em, itemsep=0em, topsep=0em]
    \item In \textbf{Sec. \ref{sec:x-1}}, we describe the process of collecting and pre-processing of the videos from online sources, then specify the label definitions for our curated datasets using audio diarization, especially the detailed explanation for our Full Video dataset. 
    \item In \textbf{Sec. \ref{sec:x-2}}, we present the architectural details of our MM-When2Speak, the pretraining stages for the model, and the inference process for Full-Videos data.
    \item In \textbf{Sec. \ref{sec:x-3}}, we provide the prompt template we use to conduct response type predictions, and conduct an ablation for different methods across runs and templates to verify prompt consistency.
    \item In \textbf{Sec. \ref{sec:x-4}}, we display some sample outputs along with success/failure predictions case studies. 
    \item In \textbf{Sec. \ref{sec:x-5}}, we conduct an experiment of extending our work to predicting ``what to speak'' using different integration strategies.
    \item In \textbf{Sec. \ref{sec:x-6}}, we discuss the comparison with temporal/event-boundary methods and relation to turn-taking prediction methods, providing insights of our design.
\end{itemize}
\section{Dataset Specifications} \label{sec:x-1}

\subsection{Video Collection and Preprocessing} \label{sec:x-1.1}

Our research mainly focuses on dyadic conversation scenarios, which have cameras pointing at the faces of two participants at relatively the same height level.
In these videos, two conversational participants are separated into split-screen frames, with one shown on the left and the other on the right (or up and down, respectively), engaging in turn-taking discussion of the topics while facing to the cameras. We construct a script to download this kind of videos using \cite{ytdlp2023github}, and gather them together forming a raw dataset with 2,403 videos collected.

The objective of data collection is to collect conversational clips that contain \textbf{clean, consistent, and high-quality} visual and audio data. However, not all downloaded videos are usable since a lot of them contain extreme low-quality factors, such as long-time black screen, regular scene change, noise interferences, or severe sound overlaps, which can not be used for conversational interactions. Therefore, we conduct further analysis of all downloaded videos to filter out those with such unbearable situations. We manually review each video to examine if any of the above factors exist in each video, and eventually come up with \textbf{377 high-quality videos}.

It is worth noting that the 377 collected videos exhibit significant variation in duration, ranging from \textbf{1 minute 54 seconds} to \textbf{9 minutes 43 seconds}. Moreover, \textbf{not all segments within each video are suitable for dyadic interaction analysis}. For instance, some videos have one speaker speaking for too long, lacking necessary interactions between two conversational participants. To ensure the dataset focuses exclusively on actual dyadic conversational interactions, we manually crop each video and only obtain the interactive part. The resulting 377 cropped videos serve as the foundation of our Short Clips dataset, whose \textbf{average time is 2 minutes 13 seconds}.

\subsection{Data Annotation} \label{sec:x-1.2}

For the video modality data, we crop out the \textbf{rectangular face area} of each speaker in each video first. For audio and text modalities, we employ \textbf{audio diarization} to facilitate the annotation. That is, we extract the audio of each high-quality video, and diarize the audio into two split audios by speakers using \cite{asteroid2023hub,pyannote2024hub}, where each split audio represents only one speaker speaking, along with the timestamps indicating whenever they start and stop speaking. We then adopt Whisper \cite{whisper2023hub} to \textbf{transcribe} the diarized audio into texts with timestamps grouped by speakers. Based on the timestamps, we manage to define the three basic categories (i.e., \texttt{reaction}, \texttt{full\_response}, and \texttt{silence}), where:

\begin{figure}[t]
    \centering
    \includegraphics[width=\linewidth,keepaspectratio]{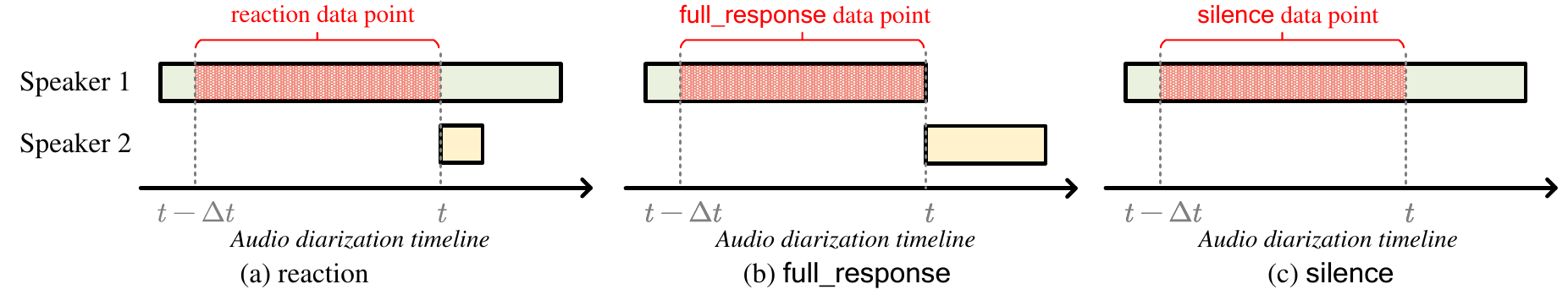}
    \caption{Illustration of short clip data point obtainment. We determine the data points by analyzing the audio diarization results. The green and yellow bars denote the diarization results of two different speakers' speaking time in one collected video. (a), (b), and (c) represent the typical data point sampling scenarios for reaction, \texttt{full\_response}, and \texttt{silence}. Note that in this illustration, when speaker 1 is speaking and does not intend to stop, speaker 2 is the listener. The \texttt{silence} data point can be sampled anywhere whenever there is only one speaker speaking. $\Delta t$ denotes window size.
    }
    \label{fig:scds}
\end{figure}

\begin{enumerate}[leftmargin=1.em, itemsep=0em, topsep=0em]
    \item \texttt{reaction}: when a listener speaks for a very short time and does not stop the speaker speaking, we consider that a ``reaction'' data point, as shown in Fig. \ref{fig:scds}(a);
    \item \texttt{full\_response}: when a listener starts speaking and the speaker stops for them to speak, we consider that a ``\texttt{full\_response}'' data point, as shown in Fig. \ref{fig:scds}(b);
    \item \texttt{silence}: when the speaker continues speaking and the listener has no intention of speaking, we consider that a ``\texttt{silence}'' data point, as shown in Fig. \ref{fig:scds}(c);
\end{enumerate}

Based on the category information, we extract $\Delta t=10$s multimodal (i.e., video, speech, and text) short clips prior to the corresponding timestamp, as one data point in the Short Clips dataset. Note that we can obtain timestamps on a per-word basis, so we are able to finely extract texts with that window.

Furthermore, we proceed to classify reactions into \textbf{seven fine-grained categories}. We first gather all texts from those short clips labeled with ``reaction'', and feed them into ChatGPT \cite{chatgpt2023openai} to classify into different reaction categories, which finally results in seven different categories: \texttt{affirmation}, \texttt{gratitude}, \texttt{greeting}, \texttt{farewell}, \texttt{surprise}, \texttt{question}, and \texttt{pondering}. We then assign these fine-grained reaction labels back to those short clips, and eventually come up with nine categories for response-type prediction.

For the generation of the seven reaction categories, we first organize all texts of reaction data into a text file, with each line representing one reaction data point. The following is a sample of the text file:

\begin{tcolorbox}[colback=gray!10!white, colframe=gray!40!black, fontupper=\fontsize{7.5pt}{9pt}\selectfont\ttfamily, sharp corners, boxrule=0.5pt, arc=0pt]

Yeah, that's right.

Oh, really?

Hmm.

Wow!

Thank you.

Absolutely.

...
\end{tcolorbox}

And then, we upload this text file to ChatGPT, and use the following prompt to obtain the seven reaction categories as well as the fine-grained category for each line in the text file. The results are saved to a newly created text file, which contains both the summarized categories and the specific reaction category assigned to each line of the uploaded text file. We also \textbf{manually examine} the resulting categorization of the reaction data to ensure no explicit outlier (e.g., misclassification from \texttt{affirmation} to \texttt{question}).

\begin{tcolorbox}[
    enhanced, 
    breakable, 
    colback=gray!10!white, 
    colframe=gray!40!black, 
    fontupper=\fontsize{7.5pt}{9pt}\selectfont\ttfamily,
    sharp corners, 
    boxrule=0.5pt, 
    arc=0pt,
]

You are a linguistic analysis assistant. Your task is to analyze a list of short utterances, where each line in the uploaded text file represents a brief reaction from a speaker during a conversation. These reactions belong to different "reaction types" (e.g., Affirmation, Surprise, etc..), which you need to summarize and categorize solely based on the uploaded text file.

\vspace{0.5em}
Please perform the following tasks:

\vspace{0.5em}
1. Read all the utterances in the text file;

2. Identify and list all distinct **reaction types** found in the text, provide a brief explanation for each type, and make sure these types make sense;

3. Assign a corresponding reaction type label to each line of text;

4. Output your results to a new text file in the following format:

   - A list of identified reaction types with explanations
   
   - The original line followed by its assigned label

\vspace{0.5em}
For example, if the input file contains the following four lines (within the double quotation marks):

\vspace{0.5em}
"Yes, I agree with you.

Oh, really?

Hmm.

Are you sure?
"

\vspace{0.5em}
You should output the following in a new text file:

\vspace{0.5em}
Identified Reaction Types:

- Affirmation: indicates agreement or confirmation

- Surprise: expresses shock, amazement, or unexpectedness

- Pondering: the process of thought according to what the speaker says

- Question: expresses doubts or raising concerns

\vspace{0.5em}
Classification Results:

- Yes, I agree with you. -> Affirmation  

- Oh, really? -> Surprise  

- Hmm. -> Pondering

- Are you sure? -> Question

\vspace{0.5em}
Now, please perform the same analysis based on the uploaded text file.

\end{tcolorbox}

Based on the output from ChatGPT, we summarize seven different reaction types, which are \texttt{affirmation}, \texttt{gratitude}, \texttt{greeting}, \texttt{farewell}, \texttt{surprise}, \texttt{question}, and \texttt{pondering} respectively. Note that we have included a human study assessing the quality of such method to generate annotations for reaction data point in our main paper. The results demonstrate that differences between human efforts and this annotation pipeline is modest rather than dramatic, suggesting that this labeling approach is comparatively reliable.

\subsection{Dataset Composition}

For a more comprehensive evaluation of our model, we compose our dataset with \textbf{two different subsets}: One subset, containing sampled 357 out of the entire collected videos, is segmented into a set of $\Delta t=10s$ \textbf{short clips}, serving to train and validate the model’s ability to perform accurate response type prediction on short, fixed-length multimodal segments; The other subset was kept as \textbf{full videos}, which contain long-range full conversations between two speakers from the remaining 20 videos, which is used to evaluate how models perform under more realistic conversational settings with continuous temporal context.

\subsubsection{Short-Clips Dataset.}\label{sec:x-1_short-clips}For the 357 collected videos used for Short-Clips dataset, based on identified reaction types, we obtain 2,433 data points for \texttt{affirmation}, 820 for \texttt{gratitude}, 253 for \texttt{farewell}, 244 for \texttt{greeting}, 192 for \texttt{question}, 289 for \texttt{surprise}, and 162 for \texttt{pondering}, which construct 4,393 reaction data points altogether. We also randomly sample 2,000 \texttt{full\_response} and 2,000 \texttt{silence} data points. Then, we use a train/test ratio of 7:3 to split each response type data point (including specific reaction types) using stratified sampling, and construct the Short-Clips-Train and Short-Clips-Test datasets.

\subsubsection{Full Videos Dataset.}\label{sec:x-1_full-videos}Instead of extracting individual clips from the 357 videos to compose Short-Clips Dataset, we only locate data points of \texttt{reaction} and \texttt{full\_response} for each video in the 20-video subset following the abovementioned annotation pipeline, where the timestamps and the corresponding response types of those data points are recorded as ground truths for the video inference. In these 20 videos, we have 52 \texttt{affirmation}, 16 \texttt{gratitude}, 8 \texttt{farewell}, 10 \texttt{greeting}, 28 \texttt{surprise}, 37 \texttt{question}, and 16 \texttt{pondering}), and 98 \texttt{full\_response} data points in total. \texttt{silence} data are not annotated because they are densely distributed across each video. The average time for these videos is \textbf{2 minutes 25 seconds}. This dataset is solely used to evaluate the model's ability to predict response types in continuous conversations, where the model must perform dense predictions of what type of response should occur temporally, rather than predicting on isolated short segments. We further introduce the inference process in Sec. \ref{sec:x-2_full-video-infer}.
\section{Model Specifications} \label{sec:x-2}

\begin{figure}[!htbp]
    \centering
    \includegraphics[width=\linewidth,keepaspectratio]{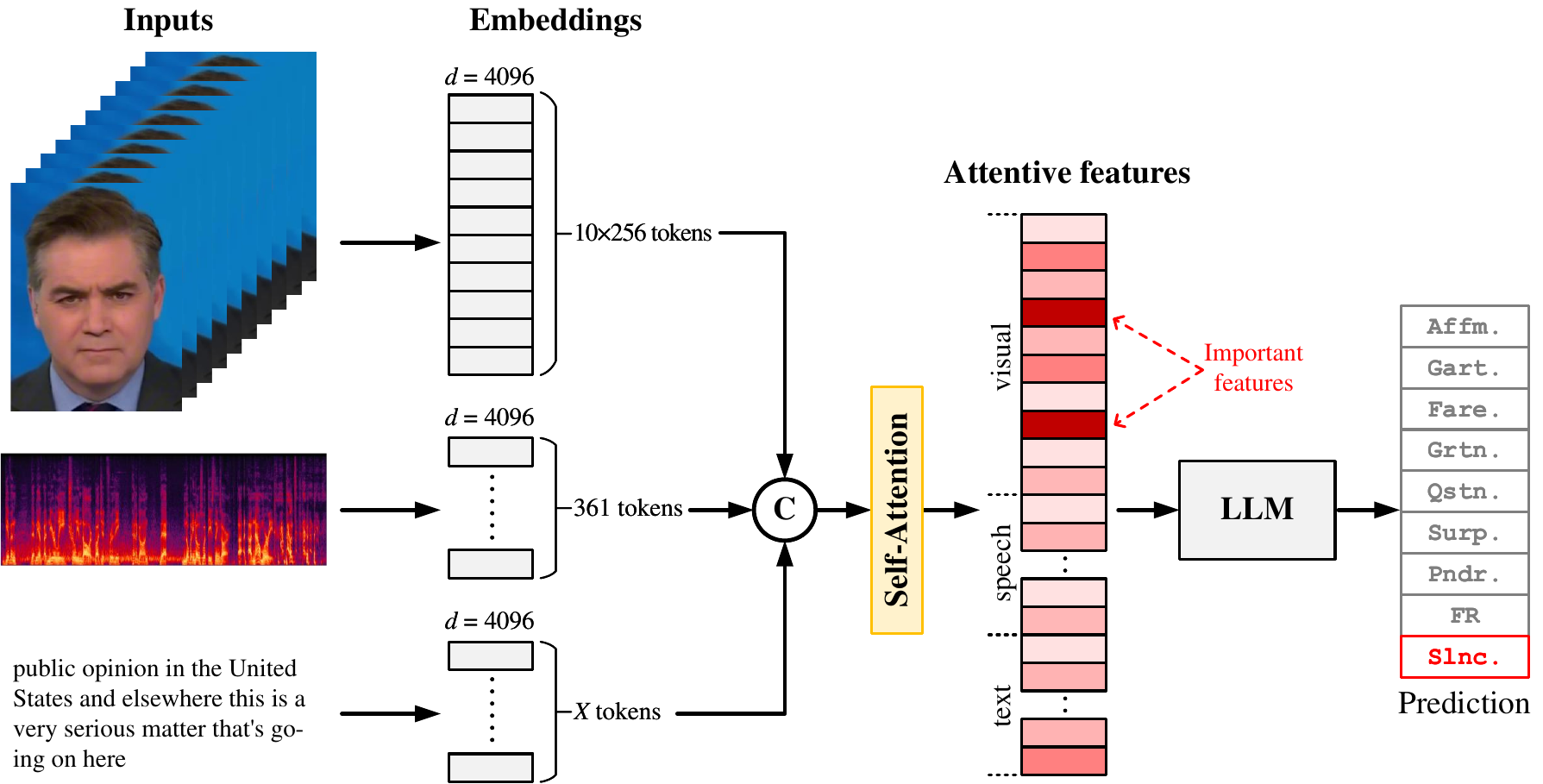}
    \caption{Feature fusion strategy for our proposed MM-When2Speak. The self-attention mechanism is used to enhance critical features in the concatenated embedding that facilitate accurate response-type predictions.}
    \vspace{-1em}
    \label{fig:mm-when2speak}
\end{figure}

\subsection{Architectural Details for MM-When2Speak}

To better model the relative importance of information across modalities, we apply a lightweight self-attention mechanism directly on the concatenated multimodal embeddings before classification. As shown in Fig. \ref{fig:mm-when2speak}, given a sequence of visual, audio, and textual embeddings extracted from the respective encoders, we first project all token embeddings into a shared latent space of 4096 dimensions. In our experiments, the visual modality consists of 10 frames, each producing 256 tokens via the visual encoder, resulting in a total of 2560 visual tokens. The audio encoder outputs 361 audio tokens for a 10-second segment, and the text encoder provides \( X \) textual tokens, depending on the utterance tokenization length. These three modalities are concatenated along the sequence dimension to form a unified sequence of (2921 + $X$) tokens, each with 4096-dimensional embeddings.

We then apply a single-layer token-wise self-attention module over this concatenated sequence to enable early-stage cross-modal interaction. This mechanism allows each token to attend to all other tokens, regardless of their modality, and dynamically reweights them based on relevance to the response-type prediction task. Unlike modality-specific fusion strategies or hard-coded weighting, this design leverages the inherent flexibility of attention to discover salient features across modalities and time. Importantly, because attention weights are computed at the token level solely based on inputs, we can infer which parts of the input contribute most to the model’s final prediction.

\begin{figure}[t]
    \centering
    \includegraphics[width=0.9\linewidth,keepaspectratio]{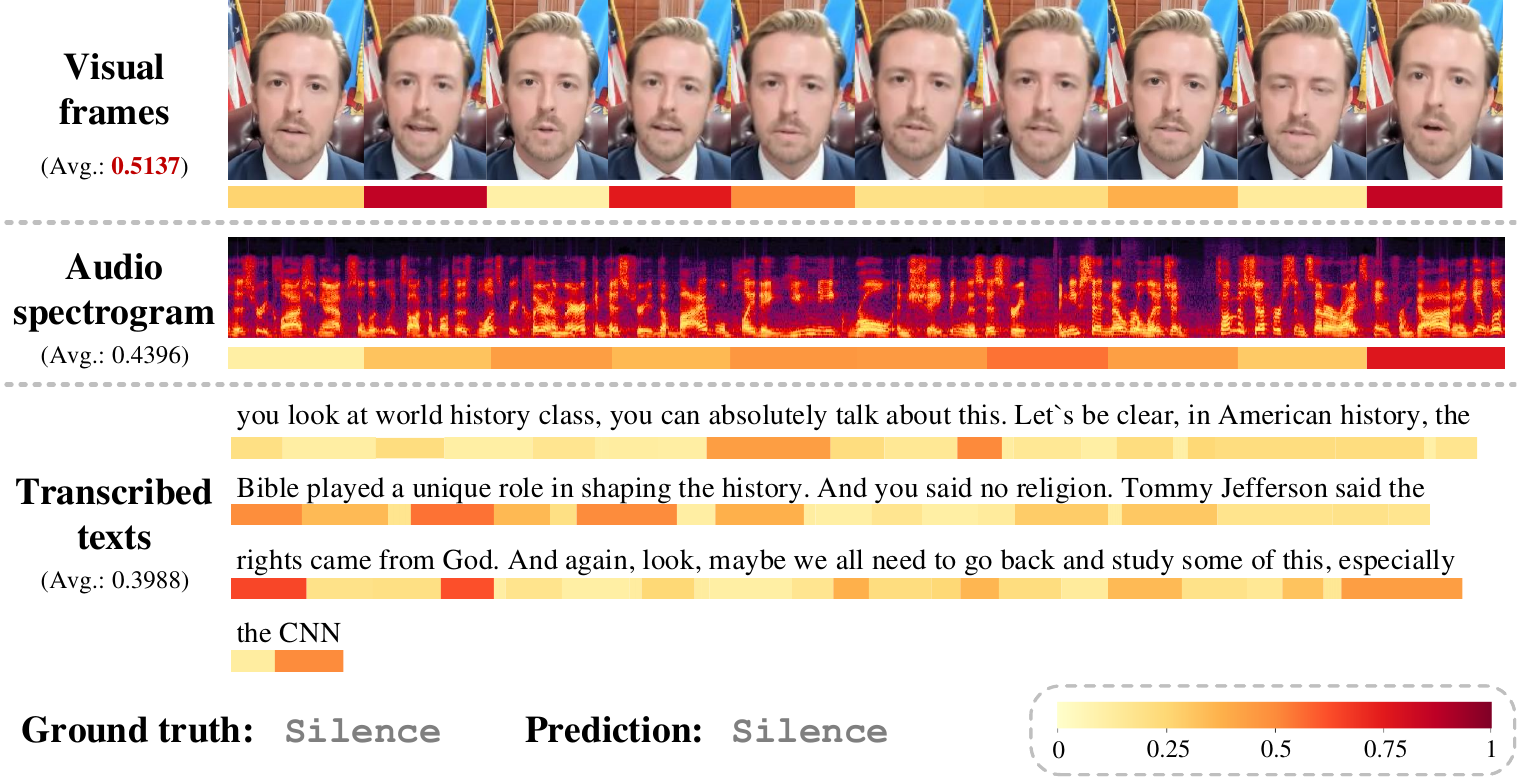}
    \caption{Visualization of token weights in the self-attention. Note that we average token weights by frames for video and speech modalities. We also report the modality-wise average attention weights, and it can be seen that the visual modality contains relatively more important information with an average attention weight of 0.5137.}
    \vspace{-1em}
    \label{fig:attn1}
\end{figure}

\begin{figure}[t]
    \centering
    \includegraphics[width=\linewidth,keepaspectratio]{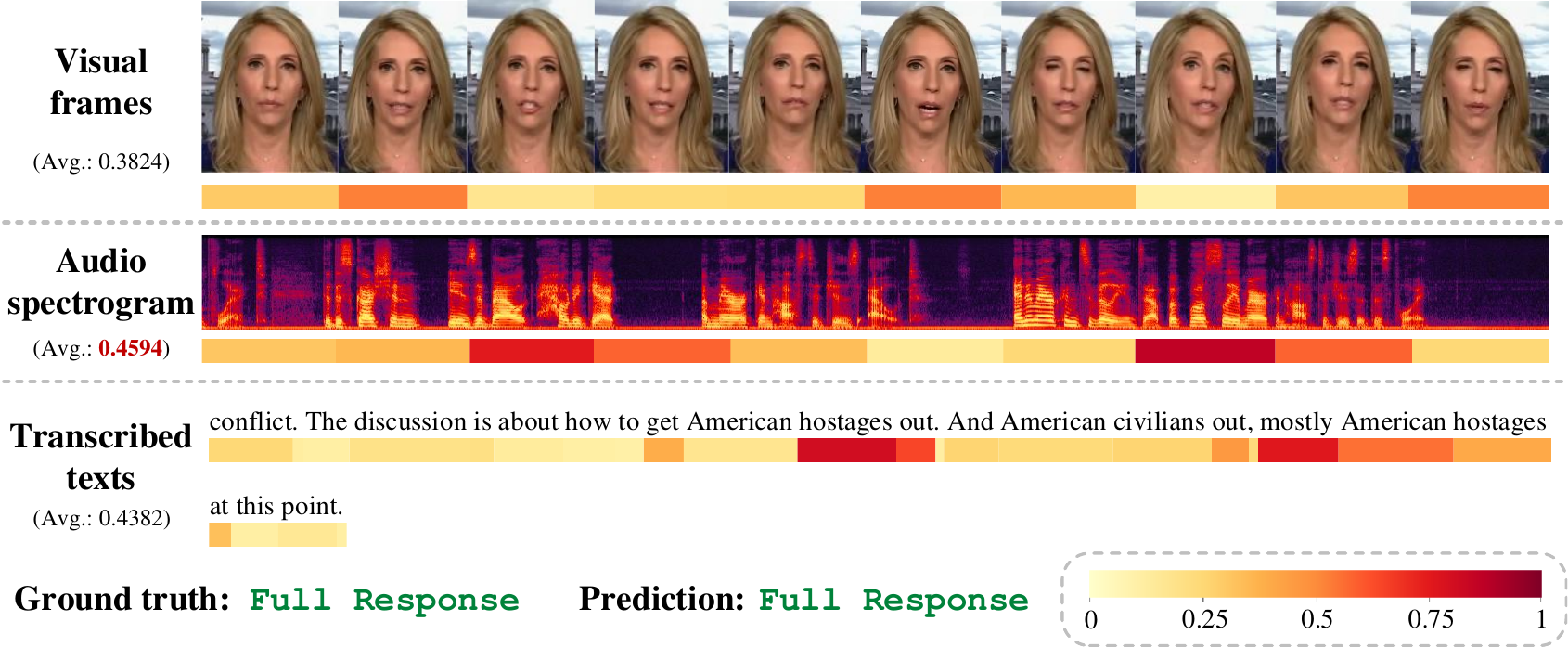}
    \vspace{-2em}
    \caption{Visualization of token weights in the self-attention. Note that we average token weights by frames for video and speech modalities. We also report the modality-wise average attention weights, and it can be seen that the speech modality contains relatively more important information with an average attention weight of 0.4594.}
    \vspace{-1em}
    \label{fig:attn2}
\end{figure}

Fig.~\ref{fig:attn1} and Fig.~\ref{fig:attn2} illustrate how the self-attention mechanism allocates weights across the concatenated sequence of multimodal inputs. 
In Fig.~\ref{fig:attn1}, the text alone does not clarify if the speaker has finished. However, the last video frame clearly shows the speaker's mouth is still open, indicating ongoing speech and that the correct response type should be ``\texttt{silence}''. As the figure illustrates, our model placed significant attention on this final video frame with an average attention weight of 0.5137 for the visual modality, resulting in the correct ``\texttt{silence}'' prediction.
On the other hand, Fig.~\ref{fig:attn2} presents an example where the visual modality is ambiguous regarding speaker's completion. Here, the audio and text modalities clearly indicate that the speaker has finished. Consequently, our model primarily attends to the audio and text modalities, leading to the correct ``\texttt{full\_Response}" prediction.

\subsection{Pretraining Stages}

Following \cite{fu2025vita}, we adopts the three-stage pretraining strategy for multimodal alignment and capability acquisition. 

\begin{enumerate}[leftmargin=2.em, itemsep=0em, topsep=0em]
    \item In the first stage, the model is trained to align visual representations with the LLM through large-scale image and video captioning data. The training procedure consists of three steps:
        \begin{itemize}[leftmargin=1.em, itemsep=0em, topsep=0em]
            \item Vision alignment where only the visual adapter is optimized;
            \item Vision understanding where both the visual encoder and LLM are fine-tuned using caption data;
            \item Instruction tuning with visual QA data to enable instruction-following capability for visual tasks.
        \end{itemize}

    \item The second stage introduces speech understanding ability. The audio encoder is first trained using speech–transcription pairs, followed by multimodal instruction tuning using speech QA data. This stage enables the LLM to process audio inputs while maintaining the visual-language capability obtained in Stage 1. 

    \item The third stage equips the model with speech generation ability. A neural codec model is first trained using text–speech pairs, and autoregressive and non-autoregressive speech decoders are subsequently optimized while keeping the LLM frozen.
\end{enumerate}

After completing the above pretraining stages, we introduce a lightweight multimodal fusion module for response-type prediction. Specifically, we apply a self-attention layer to enable adaptive multimodal aggregation. During this stage, the parameters of the LLM and modality encoders remain frozen, and only the self-attention fusion module is optimized. This design allows the model to learn task-specific multimodal interactions without affecting the pretrained multimodal capabilities. More details can be referred to \cite{fu2025vita}.

\subsection{Inference Process for Full-Video Data}  \label{sec:x-2_full-video-infer}

We use the sliding window to slide through the video with \textbf{a window size of $\Delta t=10s$} and \textbf{a stride of $\delta=0.5s$} during inference for full-video data, and each window outputs a set of multimodal (video+audio+text) context from the video for the model to predict a response type. The end time of the window, along with the prediction result, jointly create the circumstance for temporal predictions of long videos. Therefore, \textbf{the sliding window provides temporally overlapping context along the video timeline}, which transforms the full-video inference into a dense prediction problem.

Idealy, when the end time of the sliding window hits the timestamp of an annotated data point specified in Sec. \ref{sec:x-1_full-videos}, the model may output a response type corresponding to the ground truth. Specifically:

\begin{itemize}[leftmargin=1.em, itemsep=0em, topsep=0em]
    \item The model should predict a response type based on the windowed multimodal contexts, where the prediction should \textbf{correspond to the annotated ground truth} (i.e., a specific \texttt{reaction} or \texttt{full\_response}).
    \item If the end time (i.e., right end of the sliding window) does not align with any of the response-type annotation, the window should instead outputs a ``\texttt{silence}'' label;
\end{itemize}

However, the sliding window might not slide to the exact annotated timestamps in the videos, creating uncertainty for continuous predictions. To mitigate this task-related uncertainty and ambiguity, we use \textbf{a threshold of $\tau=250$ milliseconds to define a temporal neighborhood} of a given response-type data point at timestamp $t$ in a video. That is, when the end time of a window falls in the neighborhood $[t-\tau, t+\tau)$, we assign the ground truth label of the data point at time $t$ to the windowed clip, which serves as \textbf{a tolerance for labeling as well as prediction}. Note that for transcripts, audio diarization provides word-level timestamps. A word is included in a window if its starting timestamp falls within the window’s $[t-\Delta t,t)$ interval, \textbf{regardless of overlapping}. This ensures strict temporal consistency across video, audio, and text streams.
\section{Prompt Template for Model Evaluation} \label{sec:x-3}

\subsection{Original Template}

We use the following prompt to instruct models (e.g., ChatGPT, Gemini, Qwen, etc.) to perform a 9-way classification task on short conversational clips. Each input sample consists of a 10-second multimodal clip, including a frontal facial video of the speaker, their corresponding audio signal, and the exact transcript of spoken words. The model is asked to holistically assess the speaker’s verbal content, prosody, and nonverbal facial cues to determine what type of response, if any, should follow immediately after the segment. The possible target categories include brief reactions (i.e., \texttt{affirmation}, \texttt{gratitude}, \texttt{farewell}, \texttt{greeting}, \texttt{surprise}, \texttt{question}, and \texttt{pondering}), initiating full assistant replies (i.e., \texttt{full\_response}), or maintaining silence (i.e., \texttt{silence}). The prompt enforces \textbf{strict output formats} by requiring the model to respond with exactly one of nine predefined lowercase labels and to avoid any explanatory text. This setup ensures consistency and interpretability of the model’s predicted reaction types across diverse conversational contexts.

\begin{tcolorbox}[
    enhanced, 
    breakable, 
    colback=gray!10!white, 
    colframe=gray!40!black, 
    fontupper=\fontsize{7.5pt}{9pt}\selectfont\ttfamily, 
    sharp corners, 
    boxrule=0.5pt, 
    arc=0pt
]

You are an intelligent multimodal assistant (breifly "assistant" below). Your task is to classify each 10-second segment of user input into exactly one of nine predefined conversational reaction types. Each input consists of three synchronized modalities:

\vspace{0.5em}
- A video recording of a human speaker facing directly to the camera;

- An audio clip containing the speaker's voice during the same 10 seconds;

- A transcript of the exact words spoken by the user during that time window.

\vspace{0.5em}
Your goal is to analyze the entire 10-second input holistically, and determine what type of response, if any, should follow immediately after this segment ends. You must choose one and only one of the following nine response categories:

\vspace{0.5em}
1. affirmation: a short verbal confirmation or agreement (e.g., "yeah", "I see");

2. gratitude: a brief expression of thanks (e.g., "thanks", "appreciate it");

3. farewell: a polite ending or goodbye (e.g., "bye", "see you");

4. greeting: a short greeting or acknowledgment (e.g., "hi", "hello");

5. surprise: an exclamation of unexpectedness or amazement (e.g., "wow", "oh my god");

6. question: a short clarifying or curious reaction in the form of a question (e.g., "are you sure?");

7. pondering: a thoughtful pause or reflective sound (e.g., "hmm", "well");

8. full\_response: indicates that the user has likely completed speaking, and the assistant should begin a full verbal response;

9. silence: no response should be given; the assistant should remain silent and let the user continue speaking;

\vspace{0.5em}
The following are some important constraints you need to pay full attention to:

\vspace{0.5em}
- This is strictly a 9-class classification task.

- You must rely on all three modalities: video of the user's face, speech audio, and transcribed text.

- You must select exactly one label per input.

- Your output must contain only the label name, written in all lowercase letters, and must match exactly one of the categories listed above.

- Do not provide any explanation, reasoning, justification, or extra text.

\vspace{0.5em}
Now, given the following multimodal input, classify it into one of the nine categories above. Output only the category name.
    
\end{tcolorbox}

Generally, the LLM will output a short message of ``ready for input'' to indicate that it is ready for the input afterwards, such as:

\begin{tcolorbox}[
    enhanced, 
    breakable, 
    colback=gray!10!white, 
    colframe=gray!40!black, 
    fontupper=\fontsize{7.5pt}{9pt}\selectfont\ttfamily, 
    sharp corners, 
    boxrule=0.5pt, 
    arc=0pt
]

Understood. I will classify the input multimodal clip you provide for me from now. Now please upload the 10-second multimodal input for classification.
    
\end{tcolorbox}

\subsection{Performance Consistency across Templates and Runs}

In our experiments, we construct a prompt template for different models to understand our task and direct them to output exact output formats. To assess the consistency (sensitivity) of different methods to prompt formulation, we further construct two different prompts:

\vspace{-0.5em}
\begin{itemize}[leftmargin=1.em, itemsep=0em, topsep=0em]
    \item Variant 1: A \textbf{rephrasing-only} version of the original prompt, which preserves all constraints but expresses in a different way. This new prompt is presented as follows.

\begin{tcolorbox}[
    enhanced, 
    breakable, 
    colback=gray!10!white, 
    colframe=gray!40!black, 
    fontupper=\fontsize{7.5pt}{9pt}\selectfont\ttfamily, 
    sharp corners, 
    boxrule=0.5pt, 
    arc=0pt
]

You are an intelligent multimodal assistant (referred to as "assistant" below). Your task is to assign each 10-second segment of user input to exactly one of nine predefined conversational reaction categories. Every input sample contains three synchronized modalities:

\vspace{0.5em}
- A video clip showing a human speaker looking directly toward the camera;

- An audio recording capturing the speaker’s voice during the same 10-second interval;

- A transcript representing the exact words spoken by the user within that time window.

\vspace{0.5em}
Your objective is to evaluate the entire 10-second segment in a holistic manner, and determine what type of reaction, if any, should occur immediately after this segment finishes. You must select one and only one label from the following nine response categories:

\vspace{0.5em}

1. affirmation: a brief verbal acknowledgment or agreement (e.g., "yeah", "I see");

2. gratitude: a short expression of appreciation (e.g., "thanks", "appreciate it");

3. farewell: a polite closing statement or goodbye (e.g., "bye", "see you");

4. greeting: a simple greeting or acknowledgment (e.g., "hi", "hello");

5. surprise: an expression indicating unexpectedness or amazement (e.g., "wow", "oh my god");

6. question: a short inquisitive or clarifying reaction phrased as a question (e.g., "are you sure?");

7. pondering: a reflective pause or thinking sound (e.g., "hmm", "well");

8. full\_response: indicates that the user has likely finished speaking and the assistant should proceed with a complete verbal reply;

9. silence: no reaction should be produced; the assistant should stay silent and allow the user to continue.

\vspace{0.5em}
Please carefully follow the constraints listed below:

\vspace{0.5em}
- This task is strictly a nine-class classification problem.

- Your decision must be based on all three modalities simultaneously: the facial video, the speech audio, and the transcribed text.

- Exactly one label must be assigned for each input segment.

- The output must consist of only the label name, written entirely in lowercase letters, and it must exactly match one of the nine categories listed above.

- Do not include any explanations, reasoning, justification, or additional text.

\vspace{0.5em}
Now, given the multimodal input below, determine which of the nine categories it belongs to. Output only the category name.

\end{tcolorbox}

    \item Variant 2: A \textbf{few-shot} version of the original prompt, which preserves all constraints as well, but provides one example per class prior to predictions. This new prompt is presented as follows.

\begin{tcolorbox}[
    enhanced, 
    breakable, 
    colback=gray!10!white, 
    colframe=gray!40!black, 
    fontupper=\fontsize{7.5pt}{9pt}\selectfont\ttfamily, 
    sharp corners, 
    boxrule=0.5pt, 
    arc=0pt
]

\textcolor{gray}{... \# Previous original prompt}

\vspace{0.5em}
Now, I will give you an example for each of these classes, so you can have a better understanding.

\vspace{0.5em}
Here's an example for class 'affirmation': \textcolor{gray}{... \# A data example.}

\textcolor{gray}{... \# Same format for examples of other classes.}

\vspace{0.5em}
Now you will start predicting. Given the examples above and multimodal input below, determine which of the nine categories it belongs to. Output only the category name.
    
\end{tcolorbox}

\end{itemize}

\begin{figure}[!htbp]
    \centering
    \includegraphics[width=\linewidth,keepaspectratio]{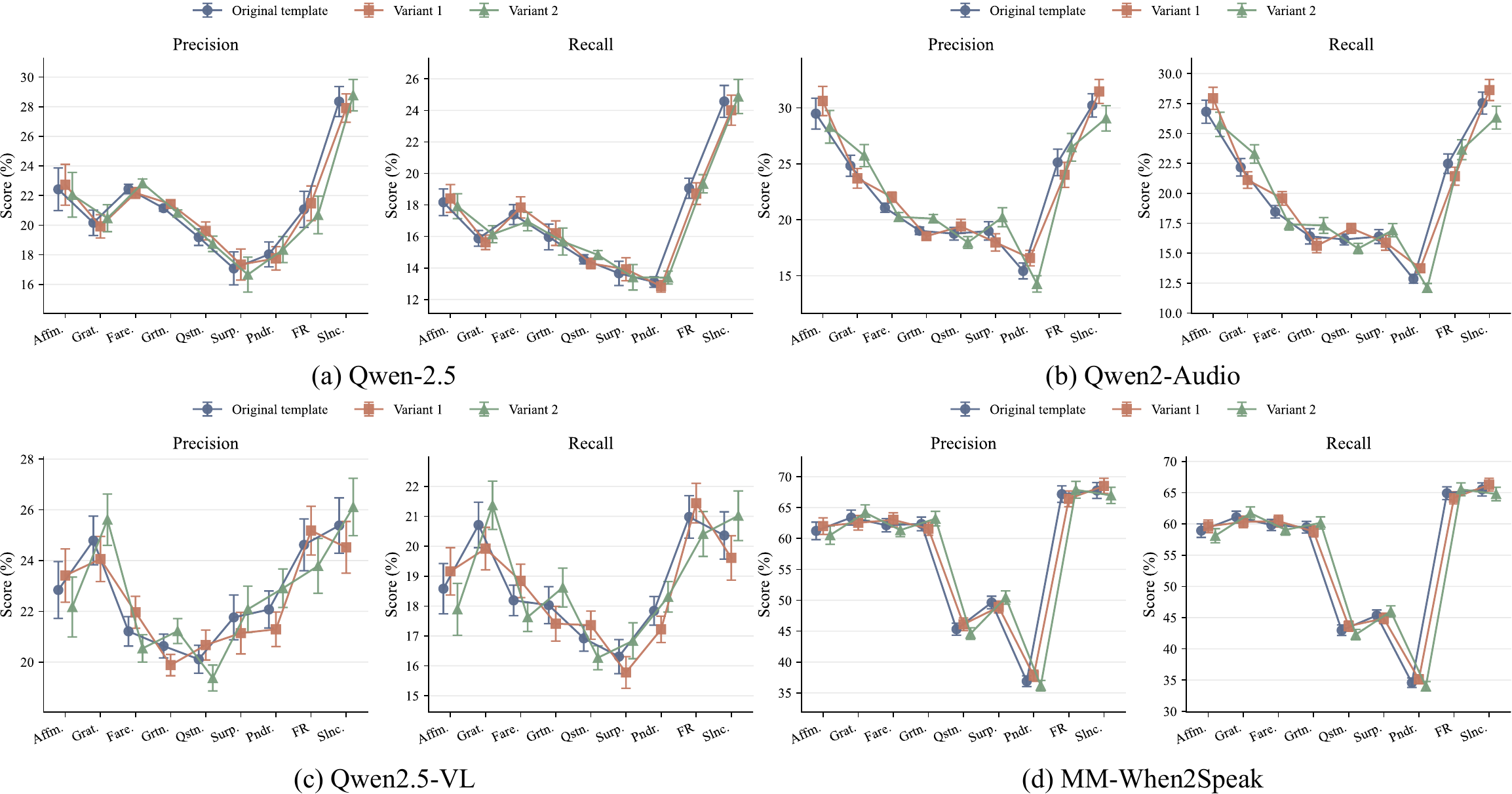}
    \caption{Point plot comparison for methods using different prompt templates for each response type. The blue/orange/green lines indicate results obtained using the original/variant 1/variant 2 prompt templates, respectively. Error bars denote the standard deviations across three runs. The results remain largely stable across different prompt templates and repeated runs, suggesting that comparison methods are relatively robust to prompt variations and indicating the reliability of the experimental findings.}
    \label{fig:ablation-templates-runs}
\end{figure}

Here we present the prediction performances of ChatGPT-5.2 (text), Qwen2-Audio (audio+text), Qwen2.5-VL (video+text), and MM-When2Speak (video+audio+text) on the Short-Clips-Test dataset. We use point plots to visualize the differences when using different prompt templates for each response type, as shown in Fig. \ref{fig:ablation-templates-runs}. The results show that the performances across different response types remain largely consistent under different prompt templates and repeated runs. The variations among prompt variants are relatively small and mostly fall within the standard deviation range. This indicates that the prediction results are not highly sensitive to prompt design, and the evaluation results for comparison methods stay consistent and robust.

\section{Representative Results and Case Study} \label{sec:x-4}

\begin{figure}[!htbp]
    \centering
    \includegraphics[width=\linewidth,keepaspectratio]{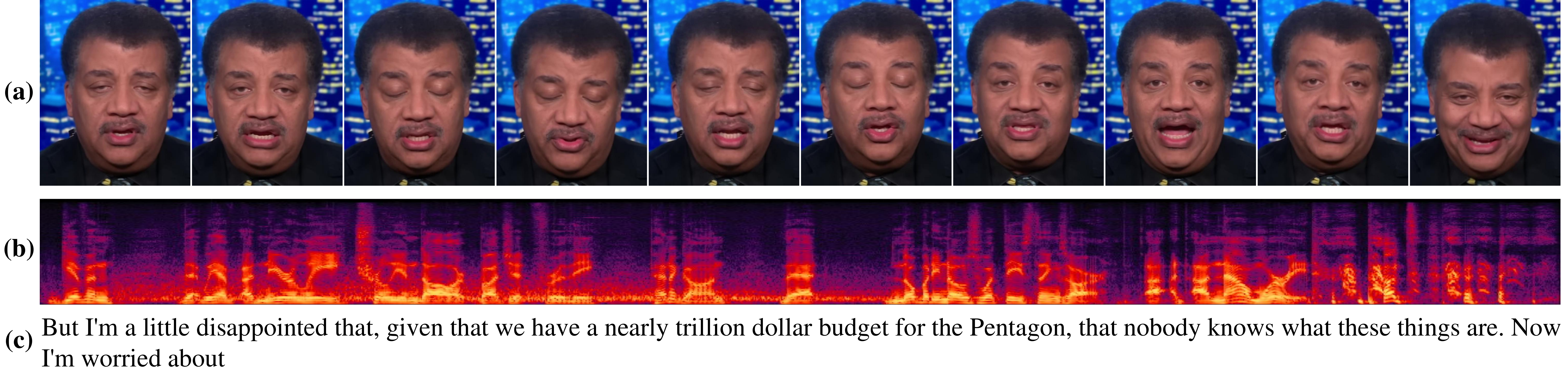}
    \caption{A sample short clip of reaction label \texttt{affirmation}. (a), (b), and (c) denote video frames, speech spectrogram, and transcribed texts, respectively.}
    \label{fig:sample-1}
\end{figure}

\begin{figure}[!htbp]
    \centering
    \includegraphics[width=\linewidth,keepaspectratio]{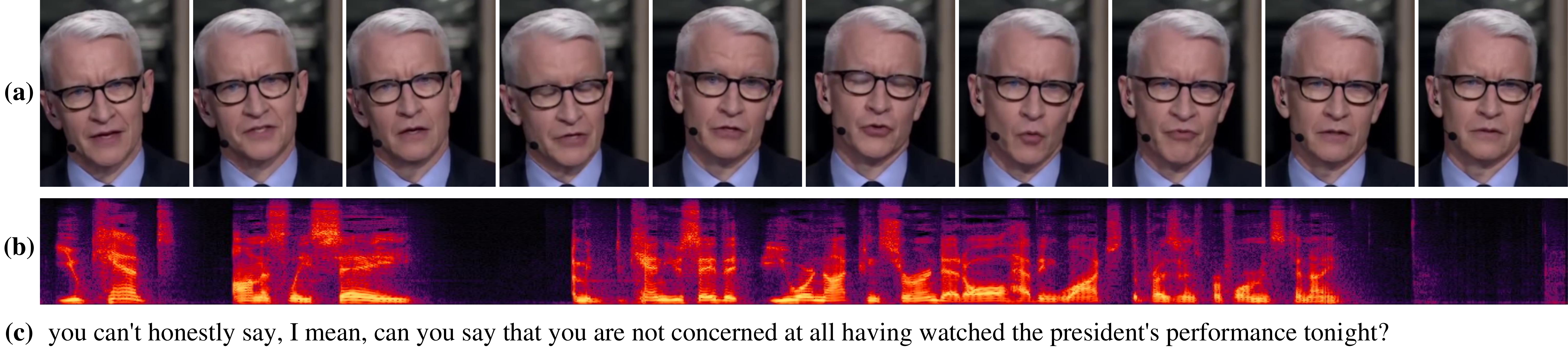}
    \caption{A sample short clip of label \texttt{full\_response}. (a), (b), and (c) denote video frames, speech spectrogram, and transcribed texts, respectively.}
    \label{fig:sample-2}
\end{figure}

\begin{figure}[!htbp]
    \centering
    \includegraphics[width=\linewidth,keepaspectratio]{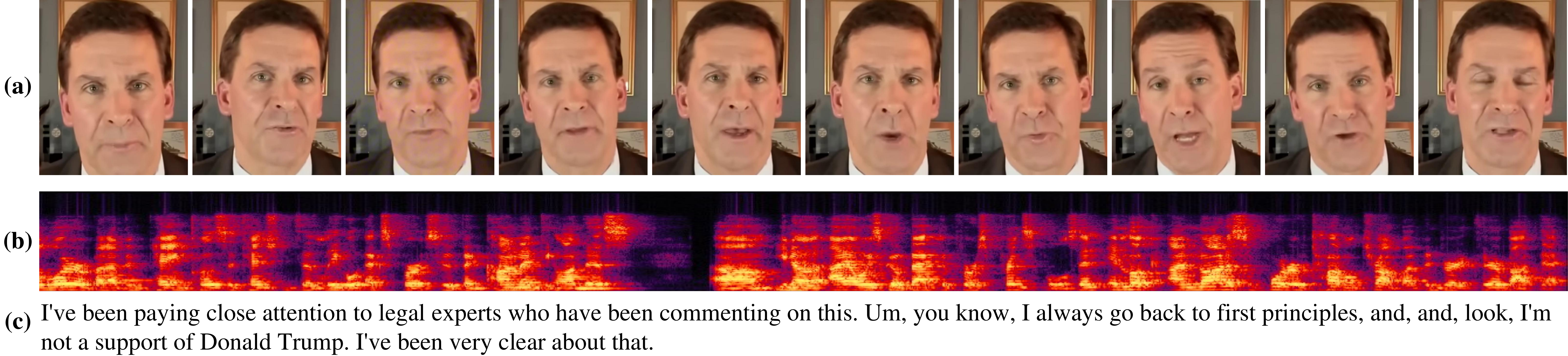}
    \caption{A sample short clip of label \texttt{silence}. (a), (b), and (c) denote video frames, speech spectrogram, and transcribed texts, respectively.}
    \label{fig:sample-3}
\end{figure}

The following are three examples of prediction of reaction (\texttt{affirmation}, as shown in Fig. \ref{fig:sample-1} and Table \ref{tab:sample-1}), \texttt{full\_response} (as shown in Fig. \ref{fig:sample-2} and Table \ref{tab:sample-2}), and \texttt{silence} (as shown in Fig. \ref{fig:sample-3} and Table \ref{tab:sample-3}), based on the prompt template described above, which is used to initialize the inference process of LLMs. We present the output of ChatGPT-5.2, Qwen2.5 for text-only modality; Qwen2-Audio for audio+text modalities; Qwen2.5-VL for video+text modalities; VITA-1.5 for video+audio+text modalities. We also include our MM-When2Speak for comparisons. Compared to other methods, the results demonstrate that our proposed MM-When2Speak predicts the response-type labels accurately and consistently.

\begin{table}[!htbp]
    \fontsize{8pt}{9.5pt}\selectfont
    \centering
    \caption{Prediction results of the Fig. \ref{fig:sample-1} multimodal input from different LLMs. The ground truth is a reaction type \texttt{affirmation}.}
    \begin{tabular}{c|c|c|c|c|c|c}
        \Xhline{1.2pt}
        \multirow{2}*{Model} & \multicolumn{2}{c|}{Text-only} & Audio + Text & Video + Text & \multicolumn{2}{c}{Video + Audio + Text} \\
        \cline{2-7}
         & ChatGPT-5.2 & Qwen2.5 & Qwen2-Audio & Qwen2.5-VL & VITA-1.5 & MM-When2Speak \\
        \hline
        Cls. & \texttt{question} & \texttt{question} & \texttt{surprise} & \texttt{surprise} & \texttt{affirmation} & \texttt{affirmation} \\
        \Xhline{1.2pt}
    \end{tabular}
    \label{tab:sample-1}
\end{table}

\begin{table}[!htbp]
    \fontsize{8pt}{9.5pt}\selectfont
    \centering
    \caption{Prediction results of the Fig. \ref{fig:sample-2} multimodal input from different LLMs. The ground truth is \texttt{full\_response} (\texttt{FR}).}
    \begin{tabular}{c|c|c|c|c|c|c}
        \Xhline{1.2pt}
        \multirow{2}*{Model} & \multicolumn{2}{c|}{Text-only} & Audio + Text & Video + Text & \multicolumn{2}{c}{Video + Audio + Text} \\
        \cline{2-7}
         & ChatGPT-5.2 & Qwen2.5 & Qwen2-Audio & Qwen2.5-VL & VITA-1.5 & MM-When2Speak \\
        \hline
        Cls. & \texttt{silence} & \texttt{silence} & \texttt{silence} & \texttt{silence} & \texttt{FR} & \texttt{FR} \\
        \Xhline{1.2pt}
    \end{tabular}
    \label{tab:sample-2}
\end{table}

\begin{table}[!htbp]
    \fontsize{8pt}{9.5pt}\selectfont
    \centering
    \caption{Prediction results of the Fig. \ref{fig:sample-3} multimodal input from different LLMs. The ground truth is \texttt{silence}.}
    \begin{tabular}{c|c|c|c|c|c|c}
        \Xhline{1.2pt}
        \multirow{2}*{Model} & \multicolumn{2}{c|}{Text-only} & Audio + Text & Video + Text & \multicolumn{2}{c}{Video + Audio + Text} \\
        \cline{2-7}
         & ChatGPT-5.2 & Qwen2.5 & Qwen2-Audio & Qwen2.5-VL & VITA-1.5 & MM-When2Speak \\
        \hline
        Cls. & \texttt{affirmation} & \texttt{affirmation} & \texttt{affirmation} & \texttt{silence} & \texttt{silence} & \texttt{silence} \\
        \Xhline{1.2pt}
    \end{tabular}
    \label{tab:sample-3}
\end{table}

Notably, natural conversational reactions are \textbf{inherently imbalanced}: high-frequency categories (e.g., \texttt{silence}, \texttt{full\_response}) dominate real interactions, while nuanced reactions (e.g., \texttt{pondering}) and boundary events (e.g., \texttt{farewell}) form the natural long tail. In our Short-Clips dataset, \texttt{affirmation} and \texttt{gratitude} account for 55.39\% and 18.66\%, whereas \texttt{question} and \texttt{pondering} are $<5\%$, and \texttt{surprise / greeting / farewell} occur in the 5--7\% range. Such skew is expected in open-domain human dialogue and correlates with lower human--human agreement. Regarding case characteristics, our qualitative analysis shows:

\vspace{0.5em}
\textbf{Successful predictions tend to occur when:}
\begin{itemize}[leftmargin=1.em, itemsep=0em, topsep=0em]
    \item Modalities reinforce each other (prosody, lexical cues, facial expressions align), e.g., positive tone + smile $\rightarrow$ \texttt{affirmation}.
    \item Visual cues are salient, e.g., widened eyes or abrupt mouth movement $\rightarrow$ \texttt{surprise / question}.
    \item Clear transition cues exist in speech, such as explicit queries or turn boundaries $\rightarrow$ \texttt{full\_response}.
\end{itemize}

\vspace{0.5em}
\textbf{Failure cases generally arise when:}
\begin{itemize}[leftmargin=1.em, itemsep=0em, topsep=0em]
    \item Signals are weak or ambiguous, such as minimal facial change or monotone speech, often causing the model to default to \texttt{silence}.
    \item Modalities contradict each other, e.g., smiling during sarcasm, causing confusion between \texttt{affirmation} and \texttt{surprise / question}.
\end{itemize}
\section{Additional Ablation Studies}  \label{sec:appendix-additional-ablations}

\subsection{Finetuned model comparisons}

Note that comparison methods in Tables \ref{tab:short-clips} and \ref{tab:full-videos} are conducted in a zero-shot manner. We further finetune some representative open-source models for a more comprehensive comparisons, and report results in Table \ref{tab:finetuned-short-clips}. Corresponding to Tables \ref{tab:short-clips} and \ref{tab:full-videos}, we can observe that not only performance improves with more modalities integrated, but also our MM-When2Speak consistently outperforms others, validating the effectiveness of our method.

\begin{table}[t]
    \fontsize{7.5pt}{6pt}\selectfont
    \setcellgapes{1pt}
    \makegapedcells
    \setlength{\tabcolsep}{1.2pt} 
    \centering
    \caption{Performance evaluations on Short-Clips-Test for finetuned models.}
    \begin{tabular}{c|c|cccccccccccccccccc}
        \Xhline{1.5pt}
        Method & Metric & \texttt{\texttt{Affm.}} && \texttt{\texttt{Grat.}} && \texttt{\texttt{Fare.}} && \texttt{\texttt{Grtn.}} && \texttt{\texttt{Qstn.}} && \texttt{\texttt{Surp.}} && \texttt{\texttt{Pndr.}} && \texttt{FR} && \texttt{\texttt{Slnc.}} \\
        \hline\hline
        \multicolumn{19}{c}{\textit{Text}} \\
        \hline
        \multirow{2}{*}{Qwen-2.5} & P & 22.50 && 18.69 && 14.98 && 15.93 && 13.13 && 12.05 && 14.79 && 21.83 && 24.10 \\
        & R & 19.38 && 15.33 && 12.50 && 10.15 && 12.91 && 8.84 && 12.05 && 19.82 && 20.45 \\
        \hline\hline
        \multicolumn{19}{c}{\textit{Audio + Text}} \\
        \hline
        \multirow{2}{*}{Qwen2-Audio} & P & 28.32 && 27.88 && 33.93 && 22.56 && 19.69 && 18.45 && 19.39 && 30.31 && 38.55 \\
        & R & 26.65 && 30.78 && 28.50 && 25.17 && 26.39 && 17.42 && 20.05 && 29.34 && 38.69 \\
        \hline\hline
        \multicolumn{19}{c}{\textit{Video + Text}} \\
        \hline
        \multirow{2}{*}{Qwen2.5-VL} & P & 33.52 && 39.14 && 40.96 && 34.82 && 27.43 && 28.38 && 26.64 && 34.52 && 45.19 \\
        & R & 30.79 && 32.74 && 36.15 && 35.57 && 23.87 && 25.66 && 27.26 && 38.67 && 40.49 \\ \hline
        \multirow{2}{*}{Video-ChatGPT \cite{maaz2024videochatgpt}} & P & 37.66 && 33.95 && 40.44 && 30.79 && 33.69 && 32.15 && 32.05 && 40.22 && 40.73 \\
        & R & 45.29 && 39.27 && 40.02 && 40.28 && 34.92 && 32.67 && 24.17 && 46.73 && 37.88 \\
        \hline\hline
        \multicolumn{19}{c}{\textit{Video + Audio + Text}} \\
        \hline
        \multirow{2}{*}{VITA-1.5} & P & 58.33 && 62.39 && 59.96 && 59.37 && 45.02 && 44.56 && 33.27 && 64.78 && 66.22 \\
        & R & 55.80 && 56.35 && 57.69 && 58.96 && 40.14 && 44.71 && 32.86 && 60.25 && 64.10 \\ \hline
        \multirow{2}{*}{MM-When2Speak}  & P & \textbf{62.21} && \textbf{64.35} && \textbf{63.15} && \textbf{63.29} && \textbf{46.26} && \textbf{50.52} && \textbf{37.78} && \textbf{68.15} && \textbf{68.78} \\
        & R & \textbf{59.86} && \textbf{61.99} && \textbf{60.79} && \textbf{60.44} && \textbf{43.91} && \textbf{46.25} && \textbf{35.45} && \textbf{65.79} && \textbf{66.42} \\
        \Xhline{1.5pt}
    \end{tabular}
    \label{tab:finetuned-short-clips}
\end{table}

\subsection{Evaluation on other datasets}

To evaluate generalizability, we conduct experiments on two other datasets, ICSI \cite{icsicorpus} (audio+text)\footnote{Dataset link: \href{https://groups.inf.ed.ac.uk/ami/icsi/}{https://groups.inf.ed.ac.uk/ami/icsi/}} and Multimediate \cite{muller2022multimediate} (video+audio+text), which are all designed for multi-party scenarios. Since our model is designed for dyadic scenarios, we adapt the dataset by arbitrarily assigning the first speaking participant as the agent side, merging the others as the user side in each session from the datasets, and processing data the same way as the Full-Videos dataset in Sec. \ref{sec:3.2}. We then conduct experiments with ChatGPT-5.2-Thinking (zero-shot), VITA-1.5 (finetuned) and MM-When2Speak (finetuned) on them in an out-of-domain (i.e., trained on our datasets and tested on ICSI/Multimediate) setting.

\begin{table}[t]
    \centering
    \fontsize{8.5pt}{7pt}\selectfont
    \setcellgapes{2pt}
    \makegapedcells
    \setlength{\tabcolsep}{2pt}
    \caption{Out-of-domain performance evaluation (Train: our dataset; test: ICSI/Multimediate).}
    \begin{tabular}{c|c|c|ccccccccc}
        \Xhline{1.2pt}
        Dataset & Model & Metric & \texttt{Affm.} & \texttt{Grat.} & \texttt{Fare.} & \texttt{Grtn.} & \texttt{Qstn.} & \texttt{Surp.} & \texttt{Pndr.} & \texttt{FR} & \texttt{Slnc.} \\
        \hline

        \multirow{6}{*}{\makecell[c]{ICSI\\(Audio+Text)}} & \multirow{2}{*}{ChatGPT-5.2} & P & 10.55 & 9.85 & 10.73 & 8.51 & 12.46 & 11.96 & 10.68 & 11.60 & 10.36 \\
         &  & R & 8.06 & 7.92 & 8.24 & 6.70 & 10.31 & 10.18 & 9.01 & 9.91 & 8.77 \\
        \cline{2-12}

         & \multirow{2}{*}{VITA-1.5} & P &  9.84 &  9.12 &  9.68 &  7.96 & 11.68 & 11.12 &  9.92 & 11.04 &  9.68 \\
         &  & R &  7.28 &  7.16 &  7.24 &  5.92 &  9.68 &  9.24 &  8.16 &  9.04 &  8.04 \\
        \cline{2-12}

         & \multirow{2}{*}{Ours} & P & \textbf{11.08} & \textbf{10.56} & \textbf{11.02} &  \textbf{9.36} & \textbf{12.88} & \textbf{12.68} & \textbf{11.40} & \textbf{12.44} & \textbf{11.16} \\
         &  & R &  \textbf{8.76} &  \textbf{8.56} &  \textbf{8.74} & \textbf{7.64} & \textbf{10.84} & \textbf{10.68} & \textbf{9.64} & \textbf{10.76} &  \textbf{9.64} \\
        \hline

        \multirow{4}{*}{\makecell[c]{Multimediate\\(Video+Audio+Text)}} & \multirow{2}{*}{VITA-1.5} & P & 13.54 & 12.49 & 14.27 & 10.98 & 13.50 & 13.16 & 11.97 & 16.27 & 16.58 \\
         &  & R & 11.18 & 10.27 & 11.81 &  8.78 & 11.30 & 10.61 &  9.54 & 14.05 & 14.73 \\
        \cline{2-12}

         & \multirow{2}{*}{Ours} & P & \textbf{15.50} & \textbf{14.45} & \textbf{16.10} & \textbf{12.66} & \textbf{14.98} & \textbf{14.73} & \textbf{13.56} & \textbf{18.45} & \textbf{18.30} \\
         &  & R & \textbf{12.97} & \textbf{11.94} & \textbf{13.50} & \textbf{10.19} & \textbf{12.69} & \textbf{12.08} & \textbf{10.99} & \textbf{16.10} & \textbf{16.28} \\
        \Xhline{1.2pt}

    \end{tabular}
    \vspace{-1.5em}
    \label{tab:out_domain_eval}
\end{table}

We report the results in Table \ref{tab:out_domain_eval}, which shows that MM-When2Speak generalizes beyond our dyadic dataset and remains effective under multi-party-to-dyadic adaptation. When trained only on our data and evaluated on ICSI/Multimediate, MM-When2Speak exhibits a consistently better performance than others, demonstrating stronger cross-domain robustness. We also observe that performance shifts likely stem from (i) ambiguity introduced by the multi-party $\rightarrow$ dyadic conversion and (ii) visual domain gaps (e.g., non-frontal faces in Multimediate), which can amplify class imbalance and boundary uncertainty. Nevertheless, MM-When2Speak remains consistently stronger than all baselines, indicating improved robustness under cross-domain conditions.

\subsection{Robustness Evaluation to Noisy Inputs}

We evaluate robustness to input degradations on Short-Clips by introducing two noise sources: \textit{audio noise} and \textit{downgraded images}. For audio, we add scaled Gaussian noise to the input waveform; for vision, we downsample and then upsample frames by a factor of 4, while keeping other modalities unchanged. As shown in Table~\ref{tab:degradation}, both MM-When2Speak and VITA-1.5 degrade under noisy conditions, but MM-When2Speak exhibits a smaller performance drop, indicating stronger robustness to input noise.

\begin{table}[t]
    \centering
    \fontsize{8.5pt}{7pt}\selectfont
    \setcellgapes{2pt}
    \makegapedcells
    \setlength{\tabcolsep}{2pt}
    \caption{Performance comparison between our method and VITA-1.5 with different degradations.}
    \begin{tabular}{c|c|c|ccccccccc}
        \Xhline{1.2pt}
        Model & Degradation & Metric & \texttt{\texttt{Affm.}} & \texttt{\texttt{Grat.}} & \texttt{\texttt{Fare.}} & \texttt{\texttt{Grtn.}} & \texttt{\texttt{Qstn.}} & \texttt{\texttt{Surp.}} & \texttt{\texttt{Pndr.}} & \texttt{FR} & \texttt{\texttt{Slnc.}} \\
        \hline\hline
        \multirow{6}*{\makecell[c]{MM-When-\\2Speak}} & \multirow{2}*{\makecell[c]{-}} & P & \textbf{62.21} & \textbf{64.35} & \textbf{63.15} & \textbf{63.29} & \textbf{46.26} & \textbf{50.52} & \textbf{37.78} & \textbf{68.15} & \textbf{58.78} \\
         &  & R & \textbf{59.86} & \textbf{61.99} & \textbf{60.79} & \textbf{60.44} & \textbf{43.91} & \textbf{46.25} & \textbf{35.45} & \textbf{65.79} & \textbf{66.42} \\
        \cline{2-12}
         & \multirow{2}*{\makecell[c]{+Audio\\Noise}} & P & 58.74 & 61.50 & 60.74 & 62.96 & 45.67 & 48.84 & 37.66 & 64.28 & 55.99 \\
         &  & R & 55.16 & 57.08 & 59.33 & 55.87 & 40.28 & 42.95 & 34.49 & 62.73 & 64.89 \\
        \cline{2-12}
         & \multirow{2}*{\makecell[c]{+Downgraded\\Images}} & P & 58.74 & 61.50 & 60.74 & 62.96 & 45.67 & 48.84 & 37.66 & 64.28 & 55.99 \\
         &  & R & 55.16 & 57.08 & 59.33 & 55.87 & 40.28 & 42.95 & 34.49 & 62.73 & 64.89 \\
        \hline\hline
        \multirow{6}*{\makecell[c]{VITA-1.5}} & \multirow{2}*{\makecell[c]{-}} & P & 38.03 & 39.68 & 39.16 & 38.81 & 28.46 & 31.23 & 23.32 & 31.36 & 35.32 \\
         &  & R & 35.14 & 36.79 & 36.27 & 35.92 & 25.61 & 28.37 & 20.50 & 28.50 & 32.44 \\
        \cline{2-12}
         & \multirow{2}*{\makecell[c]{+Audio\\Noise}} & P & 31.17 & 30.95 & 32.96 & 33.02 & 22.90 & 27.97 & 22.08 & 26.76 & 32.41 \\
         &  & R & 28.46 & 35.23 & 28.57 & 26.89 & 24.16 & 21.44 & 16.66 & 23.48 & 30.63 \\
        \cline{2-12}
         & \multirow{2}*{\makecell[c]{+Downgraded\\Visual}} & P & 31.89 & 28.16 & 31.85 & 33.15 & 23.19 & 26.52 & 20.19 & 25.14 & 33.96 \\
         &  & R & 29.17 & 33.43 & 26.68 & 25.25 & 20.08 & 26.75 & 19.62 & 21.33 & 28.69 \\
        \Xhline{1.2pt}
    \end{tabular}
    \vspace{-0.5em}
    \label{tab:degradation}
\end{table}

\subsection{Comparison with a Backchannel Detection Method}

We compare MM-When2Speak with the open-source transformer-based backchannel detector, One Stream \cite{amer2023backchannel}, on our Short-Clips reaction set. Following its official implementation, we cast backchannel detection as a \textbf{binary} task by mapping our labels to \textit{reaction} (all reaction types) vs. \textit{no reaction} (\texttt{full\_response} + \texttt{silence}). We keep the same 7:3 train/test split and report binary accuracy using One Stream's protocol (Table~\ref{tab:backchannel}). In addition, we compute per-reaction-type recall by tracing each binary prediction back to its original reaction label. MM-When2Speak outperforms One Stream by 10.29\% in binary accuracy and achieves consistently higher recall across all reaction categories, indicating strong effectiveness even under the binary backchannel setting.

\begin{table}[t]
    \centering
    \fontsize{8pt}{6.5pt}\selectfont
    \setcellgapes{1.5pt}
    \makegapedcells
    \setlength{\tabcolsep}{1.5pt}
    \caption{Backchannel detection performance comparison between One Stream and our MM-When2Speak. The ``Multi-class Cls. Recall'' denote the recall for each reaction type by dissecting the binary classification results.}
    \begin{tabular}{c|c|ccccccc}
        \Xhline{1.2pt}
        \multirow{2}*{Method} & \multirow{2}*{\makecell[c]{Binary Cls.\\Accuracy}} & \multicolumn{7}{c}{Multi-class Cls. Recall} \\
        \cline{3-9}
         &  & \texttt{\texttt{Affm.}} & \texttt{\texttt{Grat.}} & \texttt{\texttt{Fare.}} & \texttt{\texttt{Grtn.}} & \texttt{\texttt{Qstn.}} & \texttt{\texttt{Surp.}} & \texttt{\texttt{Pndr.}} \\
        \hline
        \makecell[c]{One Stream} & 48.32 & 56.33 & 53.63 & 48.17 & 52.09 & 30.66 & 37.12 & 29.68 \\
        MM-When2Speak & \textbf{58.61} & \textbf{59.86} & \textbf{61.99} & \textbf{60.79} & \textbf{60.44} & \textbf{43.91} & \textbf{46.25} & \textbf{35.45} \\
        \Xhline{1.2pt}
    \end{tabular}
    \vspace{-1.5em}
    \label{tab:backchannel}
\end{table}

\subsection{Transferability for multimodal strategy}

We switch the default LLM backbone from Qwen2-7B-base used in our MM-When2Speak to Qwen2-7B-instruct and Qwen2.5-7B-instruct for evaluating the transferability of our proposed multimodal fusion strategy. We follow the original pretraining pipeline and retrain the model, and report the performance on Short-Clips-Test in Table \ref{tab:ablation-llm}. Results show that all three different LLM backbones achieve high performances, suggesting that our proposed multimodal strategy is transferable and can be generalized to different LLM backbones.

\begin{table}[t]
    \fontsize{8pt}{6.5pt}\selectfont
    \setcellgapes{1.5pt}
    \makegapedcells
    \setlength{\tabcolsep}{1.5pt}
    \centering
    \caption{Transferability of our method on multimodal strategy with different LLMs.}
    \begin{tabular}{c|c|cccccccccccccccccc}
        \Xhline{1.5pt}
        Base Model & Metric & \texttt{\texttt{Affm.}} && \texttt{\texttt{Grat.}} && \texttt{\texttt{Fare.}} && \texttt{\texttt{Grtn.}} && \texttt{\texttt{Qstn.}} && \texttt{\texttt{Surp.}} && \texttt{\texttt{Pndr.}} && \texttt{FR} && \texttt{\texttt{Slnc.}} \\
        \hline
        
        \multirow{2}{*}{Qwen2-7B-base} & P & 62.21 && 64.35 && 63.15 && 63.29 && 46.26 && 50.52 && 37.78 && 68.15 && 68.78 \\
         & R & 59.86 && 61.99 && 60.79 && 60.44 && 43.91 && 46.52 && 35.45 && 65.79 && 66.42 \\
        \hline
        
        \multirow{2}{*}{Qwen2-7B-instruct} & P & 63.29 && 66.98 && 63.78 && 62.19 && 50.76 && 54.69 && 39.02 && 67.83 && 69.95 \\
         & R & 59.97 && 63.29 && 62.01 && 58.48 && 48.39 && 47.03 && 35.79 && 65.48 && 68.28 \\
        \hline
        
        \multirow{2}{*}{Qwen2.5-7B-instruct} & P & 66.86 && 65.39 && 64.80 && 64.71 && 51.05 && 53.89 && 44.07 && 62.78 && 68.99 \\
        & R & 61.16 && 64.64 && 62.29 && 50.54 && 49.43 && 49.82 && 37.38 && 65.55 && 67.03 \\ 
        \Xhline{1.2pt}
    \end{tabular}
    \vspace{-1em}
    \label{tab:ablation-llm}
\end{table}
\section{Extending to ``What to Speak''} \label{sec:x-5}

It is noteworthy that our work is designed to \textbf{decouple fine-grained response timing/type prediction from response content generation}. This separation is deliberate: ``when to speak'' provides a \textbf{low-level conversational control signal} that determines whether an agent should react and what kind of reaction (e.g., \texttt{affirmation}, \texttt{pondering}) should be produced, while ``what to speak'' corresponds to content generation. Predicting response types provides an interpretable intermediate representation that enables \textbf{controllable response behaviors and facilitates integration with different response generators}. Such a modular design simplifies the learning problem, avoids entangling heterogeneous objectives, and enables clear evaluation of response timing and types. This separation is also consistent with many practical conversational systems, such as AI assistants, customer-service agents, and video-call companions.

To further investigate how our framework can be extended from ``when to speak'' to ``what to speak'', we explore the following four possible strategies for generating response contents conditioned on the predicted response types:

\vspace{-0.5em}
\begin{itemize}[leftmargin=1.em, itemsep=0em, topsep=0em]
    \item \textbf{Option 1: Direct response generation without response-type prediction}. As a baseline, we directly prompt a multimodal LLM to generate responses given the input contexts, without explicitly predicting response types. In this setting, we use the original VITA-1.5 model and modify the prompt template to output response contents only.
    \item \textbf{Option 2: Response generation conditioned on predicted response types}. Since the LLM in MM-When2Speak (i.e., Qwen2) inherently supports text generation, we modify the prompt template so that the model first predicts the response type and then generates the corresponding response content conditioned on the predicted type.
    \item \textbf{Option 3: Delegating response generation to external LLMs}. MM-When2Speak can also serve as a response timing/type controller in a larger conversational system. In this setting, the predicted response type and multimodal context are passed to an external LLM (e.g., ChatGPT API) to generate the final response.
    \item \textbf{Option 4: Finetuning the model with response contents}. The model can be further trained with ground-truth responses in the dataset. By incorporating response texts during training, MM-When2Speak can learn to jointly predict response types and generate corresponding responses.
\end{itemize}

We present four qualitative examples in Tables \ref{tab:wts-example1}, \ref{tab:wts-example2}, \ref{tab:wts-example3}, and \ref{tab:wts-example4} to illustrate how different integration strategies produce response contents. For each table, we display the inputs (i.e., video frames, audio spectrum, and textual transcripts) and the outputs adopting four options. Since the focus of this work is response type prediction, here we only demonstrate the \textbf{feasibility} of integrating our model with different response generation mechanisms, rather than conducting a full quantitative evaluation of response quality.


\vspace{-1em}
\begin{table}[!htbp]
    \fontsize{9pt}{7pt}\selectfont
    \setcellgapes{2.5pt}
    \makegapedcells
    \setlength{\tabcolsep}{2.5pt}
    \centering
    \caption{Example of response generation with different options for label \texttt{silence}}
    \begin{tabular}{c|c|l}
        \Xhline{1.5pt}
        \multirow{3}*{\raisebox{-2\height}{Input}} & \raisebox{4\height}{Video} & \includegraphics[width=0.6\linewidth,keepaspectratio]{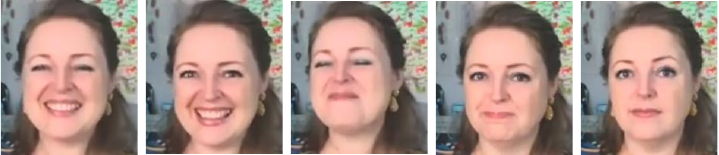} \\ \cline{2-3}
         & \raisebox{3\height}{Audio} & \includegraphics[width=0.6\linewidth,keepaspectratio]{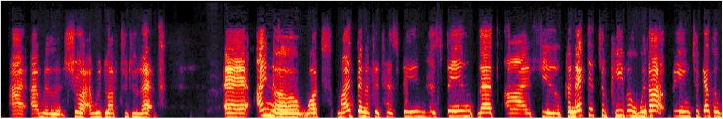} \\ \cline{2-3}
         & Text & \makecell[l]{would love to do that. I know what my benefit has been,\\ has been that I've connected to people without being toget-\\her} \\
        \hline\hline
        \multirow{4}*{\raisebox{-4\height}{Output}} & Option 1 & \makecell[l]{That's wonderful. It's amazing how meaningful connections\\ can still grow even when people aren't physically together.} \\ \cline{2-3}
         & Option 2 & \makecell[l]{\texttt{\textbf{affirmation}}: That's right.} \\ \cline{2-3}
         & Option 3 & \makecell[l]{\texttt{\textbf{affirmation}}: Wow! That sounds really nice!} \\ \cline{2-3}
         & Option 4 & \makecell[l]{\texttt{\textbf{silence}}} \\
        \Xhline{1.5pt}
    \end{tabular}
    \vspace{-0.5em}
    \label{tab:wts-example1}
\end{table}

From Table \ref{tab:wts-example1}, the ground-truth label for this contextual input is \texttt{silence}, and option 4 is the only one that correctly predicts the response type, whereas option 2 and 3 predict \texttt{affirmation} leading to incorrect generated response, and option 1 directly outputs a \texttt{full\_response} instead.

\begin{table}[!htbp]
    \fontsize{9pt}{7pt}\selectfont
    \setcellgapes{2.5pt}
    \makegapedcells
    \setlength{\tabcolsep}{2.5pt}
    \centering
    \caption{Example of response generation with different options for label \texttt{affirmation}}
    \begin{tabular}{c|c|l}
        \Xhline{1.5pt}
        \multirow{3}*{\raisebox{-2\height}{Input}} & \raisebox{4\height}{Video} & \includegraphics[width=0.6\linewidth,keepaspectratio]{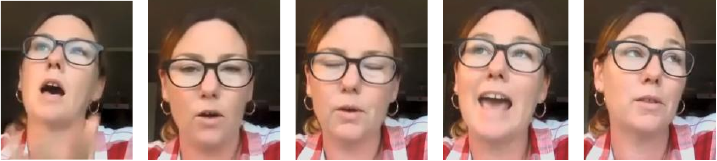} \\ \cline{2-3}
         & \raisebox{3\height}{Audio} & \includegraphics[width=0.6\linewidth,keepaspectratio]{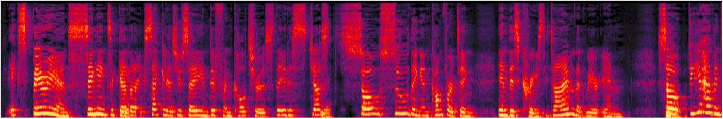} \\ \cline{2-3}
         & Text & \makecell[l]{Right, how do you, umm, balance it out, you know, ugh. I\\ think that's a great point to why it is so important teachers\\ are given the opportunity to go into each other's classroom,\\ which} \\
        \hline\hline
        \multirow{4}*{\raisebox{-4\height}{Output}} & Option 1 & \makecell[l]{Exactly! I completely agree. When teachers have opportunit-\\ies to visit each other's classrooms, it creates a great enviro-\\nment for collaboration and learning. For example, they can\\ watch each other's way to guide students as well as organize\\ the class with various activities. By observing different appr-\\oaches to instruction, teachers may gain fresh ideas for enga-\\ging students, encouraging participation, and adapting their\\ lessons to different learning styles. In the long run, this kind\\ of exchange can strengthen professional communities among\\ teachers and support continuous improvement in teaching\\ practices.} \\ \cline{2-3}
         & Option 2 & \makecell[l]{\texttt{\textbf{affirmation}}: That's right.} \\ \cline{2-3}
         & Option 3 & \makecell[l]{\texttt{\textbf{affirmation}}: Yeah, that's a really good point.} \\ \cline{2-3}
         & Option 4 & \makecell[l]{\texttt{\textbf{affirmation}}: Right.} \\
        \Xhline{1.5pt}
    \end{tabular}
    \vspace{-1em}
    \label{tab:wts-example2}
\end{table}

In Table \ref{tab:wts-example2}, the ground truth response type is \texttt{affirmation}, and options 2, 3 and 4 correctly predict this label with properly generated response contents. For option 1, it generates a lengthy paragraph of \texttt{full\_response}.

\begin{table}[!htbp]
    \fontsize{9pt}{7pt}\selectfont
    \setcellgapes{2.5pt}
    \makegapedcells
    \setlength{\tabcolsep}{2.5pt}
    \centering
    \caption{Example of response generation with different options for label \texttt{gratitude}}
    \begin{tabular}{c|c|l}
        \Xhline{1.5pt}
        \multirow{3}*{\raisebox{-2\height}{Input}} & \raisebox{4\height}{Video} & \includegraphics[width=0.6\linewidth,keepaspectratio]{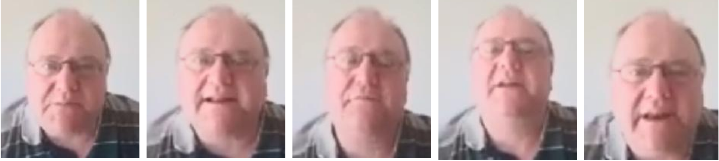} \\ \cline{2-3}
         & \raisebox{3\height}{Audio} & \includegraphics[width=0.6\linewidth,keepaspectratio]{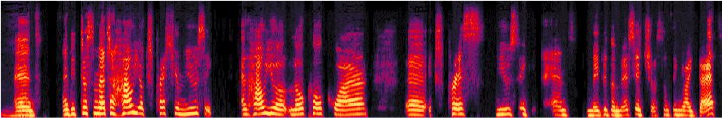} \\ \cline{2-3}
         & Text & \makecell[l]{I didn't know it was possible to feel that the, the kind of be-\\ing together, uhh, in a online media. So that was an amaz-\\ing experience for me.} \\
        \hline\hline
        \multirow{4}*{\raisebox{-4\height}{Output}} & Option 1 & \makecell[l]{I know! That's a really good point. Online spaces can some-\\times create a surprisingly strong sense of connection.} \\ \cline{2-3}
         & Option 2 & \makecell[l]{\texttt{\textbf{affirmation}}: Yeah, that's true.} \\ \cline{2-3}
         & Option 3 & \makecell[l]{\texttt{\textbf{affirmation}}: I completely agree!} \\ \cline{2-3}
         & Option 4 & \makecell[l]{\texttt{\textbf{gratitude}}: Appreciate it.} \\
        \Xhline{1.5pt}
    \end{tabular}
    \vspace{-1em}
    \label{tab:wts-example3}
\end{table}

As shown in Table \ref{tab:wts-example3}, similar to \ref{tab:wts-example1}, option 4 predicts the response type \texttt{gratitude} correctly while options 2 and 3 predict \texttt{affirmation} instead. Still, option 1 remains generating \texttt{full\_response}-like response contents.

\begin{table}[!htbp]
    \fontsize{9pt}{7pt}\selectfont
    \setcellgapes{2.5pt}
    \makegapedcells
    \setlength{\tabcolsep}{2.5pt}
    \centering
    \caption{Example of response generation with different options for label \texttt{full\_response}}
    \begin{tabular}{c|c|l}
        \Xhline{1.5pt}
        \multirow{3}*{\raisebox{-3\height}{Input}} & \raisebox{4\height}{Video} & \includegraphics[width=0.6\linewidth,keepaspectratio]{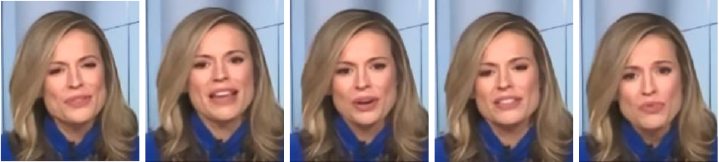} \\ \cline{2-3}
         & \raisebox{3\height}{Audio} & \includegraphics[width=0.6\linewidth,keepaspectratio]{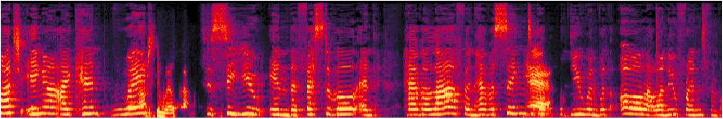} \\ \cline{2-3}
         & Text & \makecell[l]{sort of, his life lessons were, and, you know, going for so lo-\\ng, and living such a happy long life, specially his, his bea-\\utiful marriage with Rosalynn.} \\
        \hline\hline
        \multirow{4}*{\raisebox{-9\height}{Output}} & Option 1 & \makecell[l]{That's a really meaningful reflection on his life. When peop-\\le look back at such a long and fulfilling journey, the relati-\\onships they built, especially a lasting marriage like the one\\ he had with Rosalynn, often stand out as some of the most\\ powerful parts of their legacy.} \\ \cline{2-3}
         & Option 2 & \makecell[l]{\texttt{\textbf{full\_response}}: That's an important reminder that legacy isn't\\ only about public milestones. Sometimes it's the example of\\ a strong relationship and the values shown over a lifetime\\ that people remember the most.} \\ \cline{2-3}
         & Option 3 & \makecell[l]{\texttt{\textbf{full\_response}}: When people talk about a life like that, they're\\ often reflecting on the broader meaning behind it. A long ma-\\rriage, shared experiences, and a sense of dedication can leave\\ a powerful example for others.} \\ \cline{2-3}
         & Option 4 & \makecell[l]{\texttt{\textbf{full\_response}}: It's rare to see a public figure whose personal\\ and public life align in that way. The long marriage he shared\\ with Rosalynn became a powerful example of dedication and\\ partnership.} \\
        \Xhline{1.5pt}
    \end{tabular}
    \vspace{-1.5em}
    \label{tab:wts-example4}
\end{table}

From Table \ref{tab:wts-example4}, all four options generate \texttt{full\_response}-like response contents, and options 2, 3 and 4 successfully predict the \texttt{full\_response} label. 

Results from Tables \ref{tab:wts-example1} to \ref{tab:wts-example4} show that all four options are able to generate response contents under the given multimodal inputs, along with the following characteristics:
\begin{itemize}[leftmargin=1.em, itemsep=0em, topsep=0em]
    \item Option 1 directly produces full textual responses \textbf{without explicitly considering response types}, which may reduce the actual interaction across different conversational contexts.
    \item Options 2 and 3 generate short reaction-style responses \textbf{conditioned on the predicted response types}, demonstrating the ability to align responses with different conversational behaviors, although incorrect response-type predictions may occasionally lead to mismatched response contents (as in Tables \ref{tab:wts-example1} and \ref{tab:wts-example3}).
    \item Option 4 also generates responses that are \textbf{generally consistent with the contextual inputs and predicted response types}.
\end{itemize}

Overall, these examples suggest that integrating response-type prediction with response generation provides a practical way to extend our framework from ``when to speak'' to ``what to speak'' in conversational agent systems.
\section{Discussions} \label{sec:x-6}

\subsection{Differences to Temporal/Event-Boundary Methods}

Temporal/event-boundary models, such as causal streaming, CTC-style detector architectures, excel in \textbf{frame-level event segmentation} where boundaries are sharp and unambiguous (e.g., phoneme, word, or VAD decisions). These models are designed to decide \textit{exactly} when a discrete symbol starts or ends. 

In contrast, our task is \textbf{fundamentally semantic and multi-class}, with \textbf{inherently fuzzy reaction boundaries}. A reaction (e.g., \texttt{pondering}, \texttt{surprise}, \texttt{affirmation}) does not correspond to a precise frame-level onset; annotations are therefore \textbf{window-level with temporal tolerance}, not instantaneous events. Modeling must integrate multimodal cues (i.e., facial motion, prosody, lexical content) over a meaningful temporal context to output \textbf{one coherent reaction type}.

Our architecture is thus \textbf{task-aligned}: a tri-modal fusion encoder operating on a 10-second context is a natural fit for window-level semantic labels and enables clean integration into real-time sliding-window inference without altering the label space or supervision format. Adopting boundary-focused models would require \textbf{redefining labels at frame resolution}, handling overlapping or ambiguous reactions, and adding nontrivial latency and engineering complexity, while the evaluation remains window-based. We therefore view CTC/boundary architectures as \textbf{different-scope extensions}, not direct baselines for the current formulation.

\subsection{Relationship to Turn-taking Prediction}

Our task is related to prior studies on turn-taking prediction, which aim to determine when a speaker change is likely to occur in a conversation \cite{skantze2021turn,castillo2025survey}. These methods typically focus on predicting whether the current speaker will \textbf{continue speaking or take the floor} to another participant, often relying on \textbf{linguistic and prosodic cues}.

In contrast, our work focuses on predicting \textbf{listener response types} rather than speaker transitions. Instead of modeling whether a turn boundary occurs, our framework further aims to characterize the \textbf{fine-grained reactions} of a listener (e.g., affirmation, pondering, surprise) given multimodal conversational context. These reactions may occur both \textbf{within and across speaking turns} and therefore do not necessarily coincide with turn boundaries.

Moreover, while many turn-taking models primarily rely on speech signals, our framework explicitly \textbf{integrates visual, audio, and textual modalities} to capture richer conversational cues. We therefore view response type prediction as a complementary problem that \textbf{provides a finer-grained understanding} of conversational dynamics beyond turn-taking prediction.

\subsection{Limitations and Future Work}

Although our dataset covers complex dyadic interactions, real-world conversations often involve multi-party dynamics, complex role-switching, and more spontaneous or chaotic turn structures.
As a result, models trained solely on dyadic data may not generalize well to more complex, multi-person scenarios.
In addition, our annotation process (e.g., audio diarization and the categorization of reaction types) relies on empirical heuristics such as prompting ChatGPT with transcribed texts, which may introduce noise and limit labels' consistency.
Lastly, our current system focuses solely on predicting when to speak; extending it to jointly model what to speak would be a natural and valuable next step toward more fluent and human-aligned interactions with AI.

\subsection{Broader Impact}

Improving conversational AI’s timing and responsiveness through multimodal understanding can significantly enhance user experiences in applications such as virtual assistants, social robots, and accessibility tools. However, as AI becomes increasingly adept at mimicking human interactions, there is a risk of blurring the line between humans and machines. Moreover, biases in training data may result in biased AI behavior, perpetuating stereotypes or providing less effective responses for certain groups. To mitigate these risks, it is essential to ensure fairness, transparency, and an awareness of AI’s limitations while promoting responsible development. To this end, we will release our datasets and code to support further research in this area.


\clearpage
\newpage

\end{document}